\documentclass{article}

\usepackage{times}
\usepackage{graphicx} 
\usepackage{subfigure} 

\usepackage{natbib}

\usepackage{algorithm}
\usepackage{algorithmic}

\usepackage{mathtools}

\allowdisplaybreaks

\usepackage{amsmath,amssymb}
\usepackage{ mathrsfs }			

\usepackage{footmisc}			

\usepackage{color}

\usepackage{tikz}

\usepackage{amstext}               
\usepackage{amsthm}                
\usepackage{amssymb}               
\usepackage{amsopn}             
\usepackage{amsxtra}            
\usepackage{amsfonts}

\usepackage[T1]{fontenc}		
\usepackage[utf8]{inputenc}		

\usepackage{hyperref}

\usepackage[accepted]{icml2016} 

\icmltitlerunning{Neural Networks for Image Recognition}

\begin{document}

\setlength{\abovedisplayskip}{10pt}
\setlength{\belowdisplayskip}{10pt}
\setlength{\abovedisplayshortskip}{10pt}
\setlength{\belowdisplayshortskip}{10pt} 

\twocolumn[
\icmltitle{Evaluation of Neural Networks for Image Recognition Applications: Designing a $0$-$1$ MILP Model of a CNN to create adversarials}

\icmlauthor{Lucas Schelkes}{lucas.schelkes@uni-wuppertal.de}
\icmladdress{Bergische Universität Wuppertal}

\icmlkeywords{DNN, CNN, Capsules}

\vskip 0.3in
]

\begin{abstract}
Image Recognition is a central task in computer vision with applications ranging across search, robotics, self-driving cars and many others.\\
There are three purposes of this document:
\begin{enumerate}
\item We follow up on \cite{fischetti17} and show how standard convolutional neural network can be optimized to a more sophisticated capsule architecture. 
\item We introduce a MILP model based on CNN to create adversarials.
\item  We compare and evaluate each network for image recognition tasks.
\end{enumerate}
\end{abstract}
General knowledge is based on \cite{Goodfellow-et-al-2016}.

\section{Deep Neural Networks}
\label{introductionDNN}

In the following we will model a neural network in terms of a $0$-$1$ Mixed Integer Linear Program (MILP), not for training purposes, but to model well-suited instances for the network. On the contrary we can use the MILP to construct adversarial examples.\\
This chapter widely cites \cite{fischetti17}.

\subsection{Designing a MILP of DNN}
\subsubsection{Notation}
Let a Deep Neural Network (DNN) consist of $0,1, \ldots, K$ layers, where $0$ marks the input layer and $K$ identifies the output layer. Each layer $k \in \{ 0, \ldots ,K \}$ possesses $n_k$ units (or neurons) and $\text{u}(j,k)$ pinpoints the $j^{th}$ unit in layer $k$ for all $j = 1, \ldots, n_k$.\\
We assume a layered, fully connected network i.e. for all $j=1, \ldots ,n_k$ and $k=1, \ldots, K$ the unit $\text{u}(j,k)$ has $n_{k-1}$ input edges.
Each layer $k-1$ is connected to the next layer $k$ with directed edges, concretely, for $\text{u}(i,k-1)$ there exists an edge $e(\text{u}(i,k-1),\text{u}(j,k))$ that connects directly to $\text{u}(j,k)$ for all $i=1, \ldots, n_{k-1}$, for all $j= 1, \ldots, n_k$ and $k=1, \ldots, K$.
\\
The DNN can be modelled as a graph $G:=(X,E)$:\\
Let $X := \{ \text{u}(1,0), \ldots ,\text{u}(n_0,0), \ \  \ldots \ \ ,\text{u}(1,K), \ldots ,\text{u}(n_K,K) \}$ be the finite set of units in the DNN.\\
Let 
\begin{align*}
E:= \{ \\
e(\text{u}(1,0),\text{u}(1,1)),& \ldots ,e(\text{u}(1,0),\text{u}(n_1,1)), \\
e(\text{u}(2,0),\text{u}(1,1)),& \ldots ,e(\text{u}(2,0),\text{u}(n_1,1)), \ldots ,\\
e(\text{u}(n_0,0),\text{u}(1,1)),& \ldots ,e(\text{u}(n_0,0),\text{u}(n_1,1)) \\
&\vdots \\
e(\text{u}(1,K-1),\text{u}(1,K)),& \ldots ,e(\text{u}(1,K-1),\text{u}(n_K,K)), \\
e(\text{u}(2,K-1),\text{u}(1,K)),& \ldots ,e(\text{u}(2,K-1),\text{u}(n_K,K)), \ldots , \\
e(\text{u}(n_{K-1},K-1),\text{u}(1,K)),& \ldots ,e(\text{u}(n_{K-1},K-1),\text{u}(n_K,K)) \\
\}
\end{align*}
be the set of edges in the fully connected DNN.\\
\\
Let $W^{k-1}\in \mathbb{R}^{n_k \times n_{k-1}}$ be a \textit{given} weight matrix. Each weight $w \in \mathbb{R}$ is assigned to an edge $e \in E$. We define the weight $w^{k-1}_{ij}$ to be the weight between $\text{u}(i,k-1)$ and $\text{u}(j,k)$, which means to be the weight of $e(\text{u}(i,k-1),\text{u}(j,k))$. We can associate $W^{k-1} \in \mathbb{R}^{n_k \times n_{k-1}}$ to be the real-valued weights on all edges between layer $k-1$ and $k$.\\
Let $b^{k}_j \in \mathbb{R}$ be \textit{given} bias inputs for all units $\text{u}(j,k)$ with $j=1,\ldots,n_{k}$ and $k = 1, \ldots, K$. Note that $b^{k-1} \in \mathbb{R}^{n_k}$.\\
$G$ can be modelled as a directed acyclic graph, which means that for any layer $k$ there exists no edge $e(\text{u}(\cdot,k),\text{u}(\cdot,k-i))$ for all $i=1,\ldots,k$ and $k=1, \ldots, K$.\\
Let $x^k \in \mathbb{R}^{n_k}$ be the output vector of layer $k$ and $x^k_j \in \mathbb{R}$ the output value of $\text{u}(j,k)$.
Clearly $x^0$ is the input and $x^K$ the output of $G$.\\
For every layer the output $x^k$ of this layer can be calculated as follows:
\begin{align} \label{activations}
x^k := \sigma ( \underbrace{W^{k-1} x^{k-1} + b^{k-1}}_{\in \ \mathbb{R}^{n_k \times 1}}), \qquad k=1, \ldots, K
\end{align}
with the output of the previous layer $x^{k-1} \in \mathbb{R}^{n_{k-1} \times 1}$ being multiplied with the given weight matrix $W^{k-1} \in \mathbb{R}^{n_k \times n_{k-1}}$, a given bias term $b^{k-1} \in \mathbb{R}^{n_k}$ being added and this sum being fed into a pre-defined activation function $\sigma$.\\
There are several activation functions, which can be chosen from depending on the specific problem at hand. The activation function e.g. can make negative inputs to zero (like the ReLU function) or e.g. output a probability scale (like the sigmoid function). A non-linear activation function introduces non-linearity to a DNN, encouraging it to learn more complex functions. We set the activation function to be
\begin{align} \label{relu}
\sigma (x) := \mathrm{ReLU}(x) = \max \{ 0, x\} \qquad \text{with} \quad x \in \mathbb{R}.
\end{align}
ReLU is a popular choice, since it is a cheaply performed operation.
\\
Our goal is to create a MILP that modern solvers can solve - this involves \textit{not} optimizing weight parameters for training, but a model which optimizes the input $x^0$ for \textit{given} weight parameters, such that this input is the best classifiable instance. On the contrary this allows the construction of adversarial examples.\\
We need to model $\text{ReLU}(x)$ accordingly. From (\ref{activations}) we have
\begin{align} \label{relu_in_model}
x^k := \text{ReLU} ( W^{k-1} x^{k-1} + b^{k-1}),
\end{align}
in which ReLU is performed on every component of $(W^{k-1} x^{k-1} + b^{k-1}) \in \mathbb{R}^{n_k}$.
To this end, we can write the linear conditions
\begin{align} \label{linear_conditions}
&W^{k-1} \, x^{k-1} + b^{k-1} \ = \ x^{k} - s^k  \\
&\text{with} \quad x^k \geq 0, \ s^k \geq 0 \notag \\
&\forall \ k=1, \ldots, K \notag .
\end{align}
to separate the positive and the negative part.
If $W^{k-1} \, x^{k-1} + b^{k-1} \geq 0 \ \, \forall \ k$ , we choose $x^k \geq 0 \ \, \forall \ k$ thus leading to $s^k=0 \ \, \forall \ k$ and $W^{k-1} \, x^{k-1} + b^{k-1} = x^k$ (the range of $k$ is defined in (\ref{linear_conditions})).
If $W^{k-1} \, x^{k-1} + b^{k-1} \leq 0 \ \, \forall \ k$, we choose $s^k \geq 0 \ \, \forall \ k$ thus leading to $x^k=0 \ \, \forall \ k$ and $W^{k-1} \, x^{k-1} + b^{k-1} = 0 - s^k$. Due to (\ref{relu_in_model}), this implicates $W^{k-1} \, x^{k-1} + b^{k-1} = 0 \ \, \forall \ k$, which reflects the ReLU property for negative input.\\
The above solution $x^k$ and $s^k$ is not unique, since $x^k+\delta$ and $s^k+\delta$ are also solutions for any positive $\delta$ and $\forall \ k$. To achieve uniqueness, we must find a solution in which $\delta\neq0$ does not apply.
One may think to minimize $x^k+s^k \ \, \forall \ k$, but this would implicate an undesired minimisation of the ReLU-function. One may also think to introduce $x^k s^k \leq 0 \ \, \forall \ k$ as a non-linear constraint to achieve that either $x^k$ or $s^k$ will be $0 \ \, \forall \ k$, however this would contradict the MILP approach of having just linear constraints.\\
We solve this by introducing an activation variable $z^k \in \{ 0,1 \}$ to the model:
\begin{align}
&z^k = 1 \quad \rightarrow \quad x^k \leq M^{+}(1-z^k) \label{z1}\\
&z^k = 0 \quad \rightarrow \quad s^k \leq M^{-}z^k \label{z0}\\
&z^k \in \{0,1\}, \label{z01}\\
& \forall \ \, k=1, \ldots, K \notag
\end{align}
where $ 0 < M^+,M^- < \infty$ are pre-calculated bounds such that $-M^- \leq W^{k-1} \, x^{k-1} + b^{k-1} \leq M^+$ is valid for all $k = 1, \ldots, K$. These bounds are calculated by a MILP solver.\\
Conditions (\ref{z1}) - (\ref{z01}) is equivalent to
\begin{align}
&z^k = 1 \quad \rightarrow \quad x^k \leq 0 \notag \\
&z^k = 0 \quad \rightarrow \quad s^k \leq 0 \notag \\
&z^k \in \{0,1\} \notag \\
&\forall \ \, k=1, \ldots, K. \notag
\end{align}
If $z^k=1 \ \, \forall \ k$, then $x^k=0 \ \, \forall \ k$, which means that the corresponding unit in $G$ is not activated. This would encourage a trivial solution of $G$, however this is undesirable, because our goal is to find well constructed instance that are correctly classified. Consequently, we penalize the instance of $z^k=1 \ \, \forall \ k$ in the cost function to avoid a trivial solution. Incorporating the binary variable $z^k$ into the objective function qualifies our model as a $0$-$1$ Mixed Integer Linear Program.\\
In order for modern solvers to solve a MILP efficiently, we introduce upper- and lower bounds for $x^k$ and $s^k$:
\begin{align}
&lb^k_j = \overline{lb}_j^k = 0, \qquad k=1, \ldots, K \\
&ub_j^k, \overline{ub}^k_j \in \mathbb{R}_+ \cup \{+\infty\}. 
\end{align}
One way of calculating tight upper bounds is to step through all units and for every $u(j,k)$ we delete all constraints and variables associated with any other unit in either the same layer or in any higher layer, and then we solve the model (\ref{MILP_model1})-(\ref{MILP_model3}) in one round to maximize $x_j^k$ and in a second round to maximize $s_j^k$. This gives a far more accurate tight upper bounds for each units output $x_j^k$ and accelerates MILP solvers.\\
\\
Putting all pieces together delivers the following $0$-$1$ MILP model as presented by \citep{fischetti17}:

\begin{align} 
\min \quad \sum_{k=0}^K \sum_{j=1}^{n_k} c^k_j \, x^k_j \ + \ \sum_{k=1}^K \sum_{j=1}^{n_k} \gamma^k_j \, z^k_j \label{MILP_model1}
\end{align}

\begin{align} 
&\overbrace{ 
\sum_{i=1}^{n_k -1} w_{ij}^{k-1}\,  x_i^{k-1} + b_j^{k-1} \ = \ x_j^k - s_j^k }  \notag \\
&x_j^k \ , \ s_j^k \ \geq \ 0 \notag \\
&z_j^k \, = \, 1  \ \to \ x_j^k \ \leq \ 0 \notag\\ 
&z_j^k \, = \, 0 \ \to \ s_j^k \ \leq \ 0  \notag\\
&\underbrace{ z_j^k \, \in \, \{0,1\} \notag \qquad \qquad \qquad \qquad \qquad } \\
&\forall \ k=1, \ldots, K, \quad  \forall \ j=1, \ldots, n_k \label{MILP_model2}
\end{align}

\begin{align} 
&\overbrace{
lb_j^k \leq x_j^k \leq ub_j^k} \notag \\
&\underbrace{ \overline{lb}_j^k \leq s_j^k \leq \overline{ub}_j^k }   \notag \\
& \forall \ k =0, \ldots, K, \quad \forall \ j=1, \ldots, n_k \label{MILP_model3}
\end{align}

This model formulation is feasible, since for any \textit{fixed} input $x^0$ i.e. $lb_j^0 = ub_j^0 \quad \forall \ j=1 \ldots n_0$, every $x_k$ is uniquely defined by (\ref{relu_in_model}).\\
We seek minimal unit activation values $x_j^k$ in a matter such that $x_j^0 \ \, \forall \ j=1, \ldots, n_0$ is a well constructed example.\\
Surely $x_j^k=0$ for all $j=1 \ldots n_k$ and for all $k=1 \ldots K$ is a solution, but trivially superfluous. To avoid such superfluous solution to be optimal, we set $z^k$ into the cost function, so that any unit with $0$ activation value will be penalized. We seek $z$ to be $0$ as often as possible i.e. we penalize the occasion $\text{ReLU}(x)=0$. Even though the parameters for $G$ can be negative, it is only logical for a DNN to have a non-negative output. As said before, applying $\text{ReLU}(x)=0$ encourages a trivial solution for $G$ (namely $x^0=0$), so we can set each $\gamma_j^k$ to a chosen non-negative value in accordance of how greatly we want to penalize the possibility of the trivial solution problem occuring.\\
Furthermore, we can set $c^k_j := 1 \ \forall j \ \forall k$\label{costs_of_xjk}.
In this way, there are no specific units or layers of $G$ that are penalized sharper or milder.

\subsection{Creating adversarial examples}
\label{DNN:adversarial}
\begin{figure}[!htb]
\vskip 0.2in
\begin{center}
\centerline{\includegraphics[width=\columnwidth]{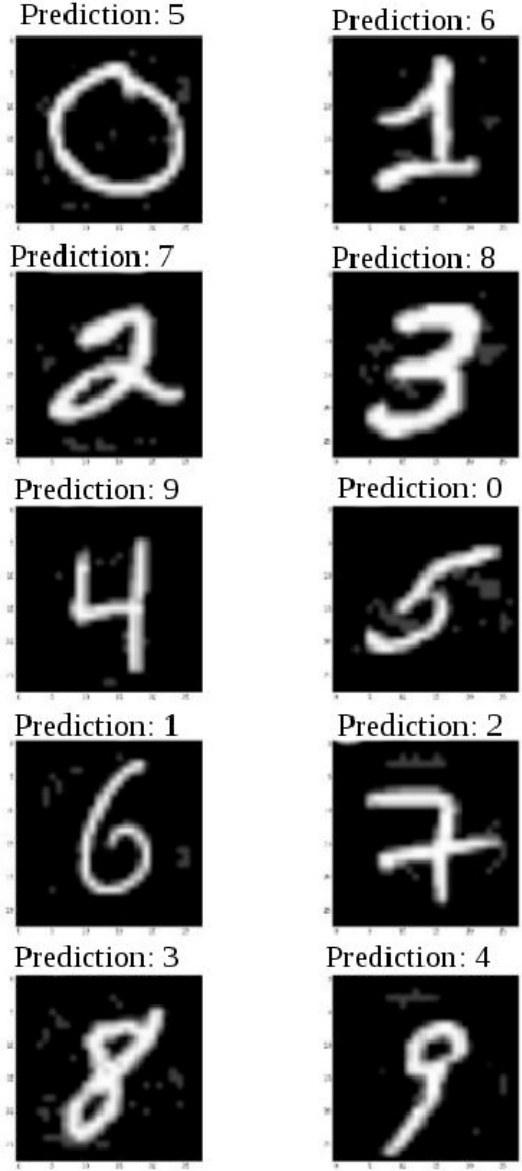}}
\caption{Adversarial instances computed by \cite{fischetti17}, including condition $\epsilon_j \leq 0.2$ for all $j=1,\ldots,n_0$. The subtle changes to the pixels are hardly recognizable to the human eye, but manage to trick the DNN.}
\label{adversarials}
\end{center}
\vskip -0.2in
\end{figure}

The described model $G$ however is not suited for training. In a DNN we have weight parameters $w_j^k$ and $b_j^k$ to be optimized, but these are fixed in $G$. We do not have any training involved in the model. Instead, $G$ is designed to implicitly compute the best possible input example $x_j^0 \ \, \forall \ j=1, \ldots, n_0$, that can best be classified by the network.\\
Inversely, we can modify $G$ to compute input examples that are worst possibly classified by the network. This will result in slightly different inputs, called adversarial examples, that the DNN will missclassify upon.\\
One application is the MNIST dataset consisting of hand-written digits as image instances.
If an image of a digit $x^0$ is classified correctly as $d$, the goal of $G$ is to find a similar image $\tilde{x}^0$ which is classified as $\tilde{d}$, with $\tilde{d}\neq d$. As \cite{fischetti17} proposes, we can set $\tilde{d} = (d+5) \mod 10$, so the adversarial image of a $3$ should have label $2$.\\
Say we want the activation of the required wrong digit in layer $K$ to be at least $20\%$ larger than any other activation, we get
\begin{align} \label{20proactivation}
\tilde{x}^K_{\tilde{d}+1} \ \geq \ 1.2 \cdot \tilde{x}_{j+1}^K \quad \forall \ j=\{0,\ldots ,9 \} \backslash \tilde{d}.
\end{align}
One can also think of modifying the cost function accordingly
\begin{align}
\min \quad \sum_{k=0}^{K-1} \sum_{j=1}^{n_k} \underbrace{c_j^k}_{=1} \,\tilde{x}_j^k \ + \ \sum_{j=0}^{9} c_{j+1}^K \, \tilde{x}_{j+1}^K,
\end{align}
with $c_{\tilde{d}+1}^K$ as negative cost: we can encourage the activation of the required wrong digit $\tilde{d}$. Conceivably, we can further penalize high activations of the other units $\tilde{x}_{j+1}^K \ \, \forall \ j=\{ 0, \ldots, 9 \} \backslash \tilde{d}$ with positive costs.\\
For the adversarial $\tilde{x}^0$ to be as similar as possible to $x^0$, we change every image pixel, such that the difference between them is close to $0$:
\begin{align} 
&\min \ \sum_{j=1}^{n_0} \epsilon_j \label{min_sum_dj} \\
-\epsilon_j \ \leq \ &x_j^0 - \tilde{x}_j^0 \ \leq \ \epsilon_j \\
&\epsilon_j \ \geq \ 0 \\
&j=1, \ldots, n_0 \label{dj_as_boundaries}
\end{align}
Adding (\ref{20proactivation}) and (\ref{dj_as_boundaries}) to $G$ modifies the model to construct adversarials $\tilde{x}^0$:

\begin{align} 
&\min \quad \Bigg(  \sum_{k=0}^{K-1} \sum_{j=1}^{n_k} \underbrace{c_j^k}_{=1} \,\tilde{x}_j^k \ + \ \sum_{j=0}^{9} c_{j+1}^K \, \tilde{x}_{j+1}^K \\
&+ \ \sum_{k=1}^K \sum_{j=1}^{n_k} \gamma^k_j \, z^k_j \ + \ \sum_{j=1}^{n_0} \epsilon_j \Bigg)
\end{align}

\begin{align} 
&\overbrace{ 
\sum_{i=1}^{n_k -1} w_{ij}^{k-1}\,  \tilde{x}_i^{k-1} + b_j^{k-1} \ = \ \tilde{x}_j^k - s_j^k }  \notag \\
&\tilde{x}_j^k \ , \ s_j^k \ \geq \ 0 \notag \\
&z_j^k \, = \, 1  \ \to \ \tilde{x}_j^k \ \leq \ 0 \notag\\ 
&z_j^k \, = \, 0 \ \to \ s_j^k \ \leq \ 0  \notag\\
&\underbrace{ z_j^k \, \in \, \{0,1\} \notag \qquad \qquad \qquad \qquad \qquad } \\
&\forall \ k=1, \ldots, K, \quad  \forall \ j=1, \ldots, n_k
\end{align}

\begin{align}
&\overbrace{
-\epsilon_j \ \leq \ x_j^0 - \tilde{x}_j^0 \ \leq \ \epsilon_j } \notag \\
&\underbrace{0.2 \ \geq \ \epsilon_j \ \geq \ 0 \qquad \qquad \qquad } \notag \\
&\forall \ j=1, \ldots, n_0
\end{align}

\begin{align}
\tilde{x}^K_{\tilde{d}+1} \ \geq \ 1.2 \cdot \tilde{x}_{j+1}^K \quad \forall \ j=\{0,\ldots ,9 \} \backslash \tilde{d}.
\end{align}

\begin{align} 
&\overbrace{
lb_j^k \leq \tilde{x}_j^k \leq ub_j^k} \notag \\
&\underbrace{ \overline{lb}_j^k \leq s_j^k \leq \overline{ub}_j^k }   \notag \\
& \forall \ k =0, \ldots, K, \quad \forall \ j=1, \ldots, n_k
\end{align}
The additional constraint $\epsilon_j \leq 0.2$ for all $j=1,\ldots,n_0$ guarantees that no pixel is changed by more than 0.2, thus leading to more pixels to be changed in total.\\
The resulting adversarial images for the MNIST dataset are presented in Figure (\ref{adversarials}) and represent a first approach to building adversarials for DNNs with MILP models.

\subsection{DNN Training}
\label{DNN_training}

The general idea of DNNs in the context of supervised learning is to feed input data into the DNN, perform a forward pass, calculate the error between the output value and the desired target value, and then update the weights using the backpropagation algorithm in order to minimize the error. By doing this iteratively many times the DNN will "learn" its weights to match the output values with the desired target values.\\
A DNNs trainable parameters are its weights $w_j^k$ and biases $b_j^k$. For training, each layer with corresponding input weights must be initialised, e.g. the \cite{he15} initialisation draws values from a Gaussian distribution centered at 0 with standard deviation = $\sqrt{\frac{2}{nb_{in}}}$, where $nb_{in}$ is the number of units of the previous layer.\\
We define the loss function as the mean squared difference of the total error: 
$$ \text{E} \, = \, \frac{1}{2\text{M}} \sum_{m=1}^M \sum^{n_K}_{i=1}  \left(\text{target}(x_i^K(m) - x_i^K(m) \right)^2$$
with $M$ equals the number of instances. The desired target activation of $u(i,K)$ of the $m^{\text{th}}$ instance is defined by $\text{target}(x_i^K(m))$. The activation value of $u(i,K)$, namely $x_k^K$, for the $m^{\text{th}}$ instance is computed by (\ref{relu_in_model}).\\
Depending on the specific problem at hand, there are different loss functions \cite{lossfunctions}.\\
A forward pass is given when applying (\ref{relu_in_model}) to every unit in the DNN.
Using the back-propagation algorithm \cite{yann89}, we can calculate how much a slight shift of an individual weight parameter affects the total error. Therefore, we calculate the derivative of $\text{E}$ in respect to $w_j^k$ and solve this with the following chain rule \cite{avrutskiy17}
\begin{align}
\notag
&\frac{\partial \, E}{\partial \, w_j^k} \, = \\
&\frac{\partial \, E(x_1^K)}{\partial \, x_1^K} \cdot \frac{\partial \, x_1^K}{\partial \, (w_1^{K-1} x_1^{K-1} + b_1^{K-1}) } \ldots \frac{\partial \, (w_j^k x_j^k + b_j^k)}{\partial \, w_j^k} \\ \notag
&+ \frac{\partial \, E(x_2^K)}{\partial \, x_2^K} \cdot \frac{\partial \, x_2^K}{\partial \, (w_2^{K-1} x_2^{K-1} + b_2^{K-1}) } \ldots \frac{\partial \, (w_j^k x_j^k + b_j^k)}{\partial \, w_j^k} \\ \notag
& \vdots \\ \notag
& + \frac{\partial \, E(x_{n_K}^K)}{\partial \, x_{n_K}^K} \cdot \frac{\partial \, x_{n_K}^K}{\partial \, (w_{n_K}^{K-1} x_{n_K}^{K-1} + b_{n_K}^{K-1}) } \ldots \frac{\partial \, (w_j^k x_j^k + b_j^k)}{\partial \, w_j^k}\\
& \forall \ k=1, \ldots , K \quad \forall \ j=1,\ldots,n_k.
\end{align}
Using a gradient descent method, all $w_j^k$ can be updated repeatedly by taking a step in the direction of steepest decrease of $E$ towards its minimum. For the sake of simplicity, we introduce batch gradient descent \cite{batch_gradient_descent}:
\begin{align} \label{gradient_descent}
w_j^k \ := \ w_j^k - \alpha \cdot \frac{\partial \, E}{\partial \, w_j^k},
\end{align}
where $\alpha$ represents a small fixed learning rate and needs to be chosen carefully. If $\alpha$ is too small, gradient descent works too slowly; if $\alpha$ is too large, gradient descent may overshoot the minimum and may fail to converge \cite{cs231:optimizers}.\\
Batch gradient descent steps through all instances of the training set, calculates the full loss function over the entire data set, and then performs one round of weight updates \cite{cs231:optimizers}. In practice the training data can have millions of instances, therefore it seems wasteful to use batch gradient descent, because training would be too slow \cite{cs231:optimizers}. Batch gradient descent has its advantages \cite{cs231:optimizers}, but more commonly other optimizers are used for more efficiency \cite{batch_gradient_descent}. Adaptive Moment Estimation (Adam) is widely used as gradient descent optimizer, as it adaptively computes the learning rate, thus sparing the need of finding an efficient learning rate experimentally \cite{batch_gradient_descent}.\\
Repeating (\ref{gradient_descent}) iteratively, until $\text{E}$ reaches its global minimum, is the process of DNN training. Local minima and sattle points are to be avoided \cite{local_minima}.

\subsection{Evaluation}

As \cite{fischetti17} shows, it is possible to use MILP models of DNNs to satisfyingly construct adversarial examples. The MILP approach, if at all, is not suited for finding optimal weight parameters.\\
We will see why DNNs are not efficient for image applications and follow up on the current state-of-the-art in image classification: Convolutional Neural Networks.

\newpage
\section{Convolutional Neural Networks}

\subsection{Introduction}

In the following, we will follow up on how Convolutional Neural Networks work, access them mathematically and finally introduce a first approach of a MILP model to create adversarial examples.

It turns out that standard neural nets are inefficient in practise for image classification. One of the main reasons for this are:
\begin{enumerate}
\item They perform weakly with highly invariant data such as different positions of the objects in the image \citep{cs231:optimizers} (see figure (\ref{dog1 image})).

\begin{figure}[!htb]
\vskip 0.2in
\begin{center}
\centerline{\includegraphics[width=80pt]{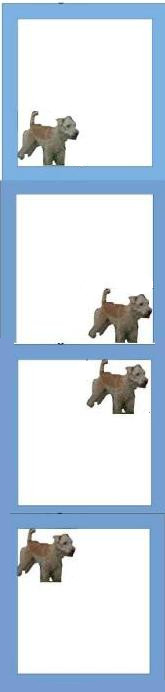}}
\caption{\cite{kaggle}\label{translational_dog}\\
A DNN, which is trained with the dog in the first image, does not recognize the dog in the other images, because it is not able to learn translational invariance. It would have to learn all of these images to recognize the same dog at different positions.}
\label{dog1 image}
\end{center}
\vskip -0.2in
\end{figure}

\medskip

\item There are extensively large amounts of network parameters to calculate e.g. an image in colour of more respectable size, e.g. 200x200x3, would mean that each neuron in the following hidden layer has $200*200*3=120,000$ weight parameters. Such great numbers of parameters increases the risk of overfitting \cite{cs231:optimizers}. 
\end{enumerate} 

Convolutional neural networks (CNNs) have proven to be the current state-of-the-art architecture for classifying images using deep learning \cite{wiki_convnets}. CNNs have the desirable property of being able to detect objects, even though they might be invariantly shifted \cite{wiki_convnets}.\\
The best so far performance marked on the MNIST database scores an error rate of $0.21\% $ \cite{wiki_mnist}, using complex CNNs and preprocessing measures.\\ 
We reflect more detailedly on how CNNs work in the following. Our goal is to introduce a mathematical approach on constructing adversarials with a MILP.\\
CNNs take advantage of image data provided in 3 dimensions: width, height and depth (number of colour channels). Concretely, CNNs consist of a sequence of layers, namely \textbf{Convolutional Layer}, \textbf{ReLu}, \textbf{Pooling Layer} and \textbf{Fully-Connected Layer} \cite{cs231:optimizers}.\\
These layers are stacked upon eachother, forming the CNN's architecture. We will see how a basic CNN architecture looks like.

\subsection{Architecture}

\subsubsection{Input layer} We define the input images as $A_\beta^1(i,j)$ for the height dimension $i=1, \ldots, h$, the width dimension $j=1, \ldots, w$ and the depth dimension $\beta = 1, \ldots, \alpha(A^1)$, where $\alpha(A^1)$ is the number of input maps (not to be confused with batch size). We assume that the next layer is convolutional (identified by superscript $1$ in paragraph (\ref{subsection:convolutional_layer}) . We think of the input images as tensors:

\begin{align} 
A_1^1 \ = \
&\begin{bmatrix} \label{input_matrix1}
A_1^1(1,1)&, \ldots , &A_1^1(1,w) \\
A_1^1(2,1)&, \ldots,   &A_1^1(2,w)\\
\vdots&  \vdots  &\vdots \\
A_1^1(h,1)&, \ldots,  &A_1^1(h,w)
\end{bmatrix}
\\ \notag \\
A_2^1 \ = \
&\begin{bmatrix} \label{input_matrix2}
 A_2^1(1,1)&, \ldots , &A_2^1(1,w) \\
A_2^1(2,1)&, \ldots,   &A_2^1(2,w)\\
\vdots&  \vdots  &\vdots \\
A_2^1(h,1)&, \ldots,  &A_2^1(h,w)
\end{bmatrix}
\\  \notag \\
&\qquad \qquad \quad \ \notag \ \, \vdots \\ \notag \\
A_{\alpha(A^1)}^1 \ = \
&\begin{bmatrix} \label{input_matrix3}
A_{\alpha(A^1)}^1(1,1)&, \ldots , &A_{\alpha(A^1)}^1(1,w) \\
A_{\alpha(A^1)}^1(2,1)&, \ldots,   &A_{\alpha(A^1)}^1(2,w)\\
\vdots&  \vdots  &\vdots \\
A_{\alpha(A^1)}^1(h,1)&, \ldots,  &A_{\alpha(A^1)}^1(h,w)
\end{bmatrix}.
\end{align}
Let $A_\beta^1(i,j) \in [0, 255]$ define the pixel integer value in the $i^\text{th}$ height, the $j^\text{th}$ width and the $\beta^\text{th}$ depth of the input image. For RGB colour images, we have $\alpha(A^1) = 3$. For these $3$ colour maps, each is a $ h \times w$ tensor with integer values between $[0,255]$, because each colour map displays its colour array as an $8$-Bit integer \cite{Goodfellow-et-al-2016}. Even though this document treats image applications i.e. $\alpha(A^1) \in \{1, 3 \}$, we will keep the depth dimension generic.

\subsubsection{Convolutional layer} \label{subsection:convolutional_layer}

\paragraph{Intuition}

\begin{figure}[!htb]
\vskip 0.2in
\begin{center}
\centerline{\includegraphics[width=\columnwidth]{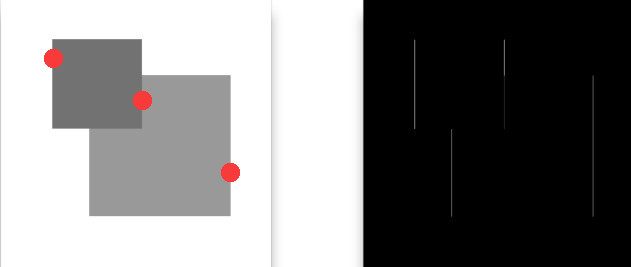}}
\caption{\cite{understanding_kernels}}\label{box_to_lines}
\end{center}
\vskip -0.2in
\end{figure}

\begin{figure}[!htb]
\vskip 0.2in
\begin{center}
\centerline{\includegraphics[width=\columnwidth]{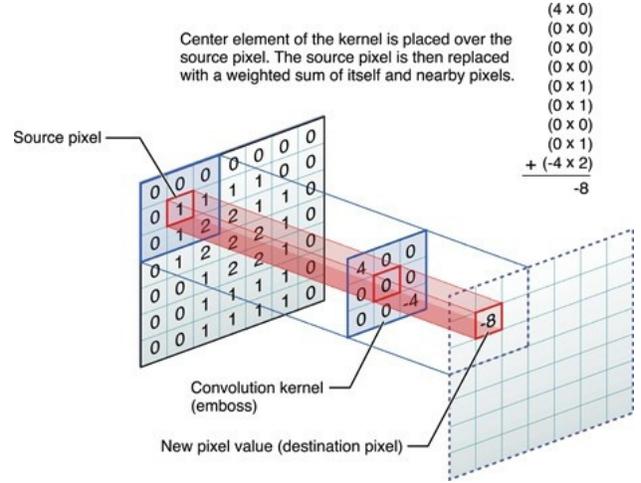}}
\caption{Illustration of convolution \cite{conv}}\label{kernel}
\end{center}
\vskip -0.2in
\end{figure}

The convolutional layer is the core element of a CNN. They consists of pre-initialised convolutional kernels of square size $f \times f$ that perform linear combinations of pixel values \cite{wiki_convnets}.\\
Applying convolutional operations to an image is the biological simulation of how the human eye works. The human visual cortex incorporates receptive fields, a cluster of neurons that result in one firing neuron. The receptive field helps the retina to identify objects in their shape and colour, adjust sharpness of vision and reduce the flood of incoming information so that signals can be processed more easily \cite{receptive_field}. Convolutional kernels imitate properties of the receptive field.
The idea is that a kernel with certain components is able to detect a certain figure in the image. According to the given components of the kernel, the kernel is able to recognize the presence of a specific figure, edge or characteristic in the image. The more accurate these components of the kernel are, the more precise it can spot that specific figure. These components can be trained, making the components the weight parameters of this layer.\\
A kernel is an operator performing dot multiplications along the pixel values according to its kernel compontents, the result is a real value (figure (\ref{kernel})). Then the kernel moves one stride further to perform the next operation. Effectively, the kernel slides through the input maps with a certain stride length \cite{cs231:optimizers}, layer by layer, over the entire width and height of the input volume, while at each position an operation is executed. This is performed on all depth maps.\\  For instance, on the left hand side figure (\ref{box_to_lines}) shows an image of two different coloured boxes. This kernel e.g. measures the differences between the borders of the two squares: 
\begin{align*}
&\begin{bmatrix}
&0 &0 &0 \\
&1 &0 &-1 \\
&0 &0 &0 \\
\end{bmatrix}
\end{align*}
\cite{understanding_kernels}

Suppose the image is given in pixel values as in (\ref{input_matrix1}). The kernel given above will calculate the pixel differences at each pixel cluster - the red dot clusters have disparate differences, as do all other clusters at the borders. The kernel outputs an integer for each border cluster and zero for all other clusters. The right hand figure (\ref{box_to_lines}) shows the output when applying this kernel. Only vertical lines are highlighted. This illustrates how this kernel works as an receptive field in order to identify vertical lines.\\
We can apply further different kernels to detect horizontal lines, edges, angles etc. The task of CNNs is to learn the values of all given kernels, such that each kernel is able to identify a certain figure in the image. Multiple convolutional layers give higher level convolutional kernels that are able to detect more comlex shapes and sophisticated figures \cite{cs231:optimizers}. The further we move up in layers, the more complex and higher level the extracted features detected by kernels get.\\
Since each kernel is moving across the input tensors, it is \textit{sharing} its weights with many input units. This makes CNN predestined to image recognition, since it reduces the amount of computation significantly. Importantly, convolutional kernels enable already learned data to be shared across space, meaning they enable translational invariance of the input data, such as figure (\ref{translational_dog}): recognizing the dog in every corner of the image, even though only one of them has been trained upon.

\paragraph{Notation}

Let $1, \ldots, C$ be the number of convolutional layers in the CNN, where $C$ is a hyperparameter to the CNNs architecture.
For each convolutional layer $c \in \{1, \ldots, C\}$ there exist $m^c \in \mathbb{N}$ kernels, where $m^c$ is also a hyperparameter. Let $k_\gamma^c$ pinpoint the $\gamma^{\text{th}}$ kernel in convolutional layer $c \in \{1, \ldots, C\}$ for all $\gamma = 1, \ldots, m^c$. For all $\gamma=1, \ldots, m^c$ and  $c \in \{1, \ldots, C\}$ we define $k_\gamma^c$ to be a tensor with square size $f^c \times f^c \in \mathbb{N} \times \mathbb{N}$:
\begin{align}
k_\gamma^c \ = \
\begin{bmatrix}
k_\gamma^c(1,1)&, \ldots, &k_\gamma^c(1,f^c) \\
k_\gamma^c(2,1)&, \ldots, &k_\gamma^c(2,f^c) \\ 
\vdots & \vdots &\vdots \\
k_\gamma^c(f^c,1)&, \ldots, &k_\gamma^c(f^c, f^c)
\end{bmatrix},
\end{align}
where $k_\gamma^c(i, j) \in \mathbb{R}$ marks the $i^{\text{th}}$ row and $j^{\text{th}}$ column component of kernel $k_\gamma^c$ with $i=1, \ldots, f^c$ and $j=1, \ldots, f^c$. Note that $f^c$ is a hyperparameter and needs to be chosen for each layer $c$.\footnote{$f^c < w^c$ and $f^c < h^c$ must hold; we assume that the kernel size is significantly smaller than the input size, since we want to have small clusters in order to accurately find small figures, lines, edges etc.}\\
Suppose that $A^c$ denotes the input of a convolutional layer $c$ and $A^c_1, \ldots, A^c_{\alpha(A^c)}$ are the input maps, where $\alpha(A^c) \in \mathbb{N}$ is the number of input maps of $c$.\\
Similarly, suppose that $B^c$ denotes the output of a convolutional layer $c$ and $B^c_1, \ldots, B^c_{m^c \cdot \alpha(A^c)}$ are the output maps. There are $m^c \cdot \alpha(A^c)$ output maps of layer $c$.\\
As \cite{uniform_distribut} suggests, one can initialize the kernels with random values drawn from a uniform distribution
\begin{align*}
k_\gamma^c \ \sim \ U \left( \pm \sqrt{\dfrac{f^c}{(\alpha(A^c) + f^c) \cdot (m^c)^2}} \right) \\
\forall \ \gamma=1, \ldots, m^c, \quad \forall \ c=1, \ldots C,
\end{align*}
where $U( \pm x )$ denotes a uniform distribution with upper and lower bounds of $\pm x$. Bias values can also be taken into account here, however we will omit these for the sake of simplicity. Each kernel can be associated with a certain form in the image that it can analyse. Therefore it is advisable to add many different convolutional filters to the layer to achieve high accuracy of identifying details \cite{uniform_distribut}.\\
The convolutional operations for every layer $c \in \{1, \ldots C \}$ are performed as follows:

\begin{align}
&\textbf{for} \quad \gamma \ = \ 1, \ldots, m^c \label{forallkernels} \\
&\qquad \textbf{for} \quad \beta \ = \ 1, \ldots, \alpha(A^c) \label{forallinputmaps}\\
&\qquad \textbf{set} \quad \delta \ := \ \beta + (\gamma - 1)\cdot \alpha(A^c) \label{foralloutputmaps}\\
&\overbrace{
B_\delta^c(1,1) \ = \ \sum_{i=1}^{f^c} \, \sum_{j=1}^{f^c} \ A_\beta^c(i,j) \cdot k_\gamma^c(i,j) \qquad \qquad \qquad \quad} \label{firstmap_firstkernel_rowone_columnone} \\
&B_\delta^c(1,2) \ = \ \sum_{i=1}^{f^c} \, \sum_{j=1}^{f^c} \ A_\beta^c(i,j+S^c) \cdot k_\gamma^c(i,j) \\
&\vdots \notag \\
&B_\delta^c \left(1, \frac{w^c-f^c+2P^c}{S^c}+1\right) \ = \notag \\ 
&\sum_{i=1}^{f^c} \, \sum_{j=1}^{f^c} \ A_\beta^c\left(i,{w^c-f^c+2P^c}+j\right) \cdot k_\gamma^c(i,j) \label{firstmap_firstkernel_rowone_columnlast} \\
&\rule{80mm}{.01pt} \notag \\
&B_\delta^c(2,1) \ = \ \sum_{i=1}^{f^c} \, \sum_{j=1}^{f^c} \ A_\beta^c(i+S^c,j) \cdot k_\gamma^c(i,j) \label{firstmap_firstkernel_rowtwo_columnone} \\
&\vdots \notag \\
&B_\delta^c \left(2, \frac{w^c-f^c+2P^c}{S^c}+1\right) \ = \notag \\ &\sum_{i=1}^{f^c} \, \sum_{j=1}^{f^c} \ A_\beta^c\left(i+S^c,{w^c-f^c+2P^c} +j\right) \cdot k_\gamma^c(i,j) \label{firstmap_firstkernel_rowtwo_columnlast}\\ 
&\rule{80mm}{.01pt} \notag \\
&\vdots \qquad \qquad \qquad \quad \qquad \qquad \vdots \qquad \qquad \qquad \qquad \quad \qquad \vdots \notag \\
&\rule{80mm}{.01pt} \notag \\
&B_\delta^c\left(\frac{h^c-f^c+2P^c}{S^c}+1,1\right) \ = \notag \\ &\sum_{i=1}^{f^c} \, \sum_{j=1}^{f^c} \ A_\beta^c(h^c-f^c+2P^c+i,j) \cdot k_\gamma^c(i,j) \label{firstmap_firstkernel_rowlast_columnone} \\
&\vdots \notag \\
&B_\delta^c\left(\frac{h^c-f^c+2P^c}{S^c}+1,\frac{h^c-f^c+2P^c}{S^c}+1\right) \ = \notag \\
&\underbrace{\sum_{i=1}^{f^c} \, \sum_{j=1}^{f^c} \ A_\beta^c(h^c-f^c+2P^c+i,w^c-f^c+2P^c +j) \cdot k_\gamma^c(i,j) \label{firstmap_firstkernel_rowlast_columnlast}} \\
&\qquad \qquad \qquad \qquad \forall \ c \ \in \ \{1, \ldots, C \} \notag
\end{align}
where $S^c \in \mathbb{N}$ is the stride step to which the kernel iterates through the rows and columns of $A_\beta^c$ and $P^c$ is the number of zero-paddings. The kernel slides through the rows and colums in steps of $S^c$; note that each layer $c$ has a fixed stride $S^c$, such that $S^c$ applies to all kernels $k_\gamma^c, \quad \gamma=1,\ldots,m^c$. Clearly $S^c \leq f^c \ \, \forall \  c$ must always be valid, otherwise the kernels would not overlap with the maps, thus losing information. Applying strides reduces the size of the output maps.
Zero-padding $P^c \in \mathbb{N}$ is the number of additional frames that are applied to $A_\beta^c$, such that a left vertical vector of zeros, a right vertical vector of zeros, a top horizontal vector of zeros and a bottom horizontal vector of zeros are appended to tensor $A_\beta^c$. This hyperparameter can be used to control the output map size.\\
Furthermore, it becomes evident that certain strides are not valid, since $\frac{w^c - f^c + 2P^c}{S^c}+1$ needs to be an integer number\footnote{same for height $h^c$ respectively.\label{h_c}} so that convolutions on only whole pixels are provided. Concretely, the constraint
\begin{equation}
(w^c - f^c) \mod S^c \ = \ 0 \quad \quad \forall \ \, c=1,\ldots, C \ \textsuperscript{\ref{h_c}} \label{stride_modulo}
\end{equation}
needs to be given for a valid stride $S^c$.
If (\ref{stride_modulo}) is not valid for a fixed $S^c$, then we can see for which $P^c$ the constraint
\begin{equation} \label{formula}
(w^c - f^c + 2P^c) \mod S^c \ = \ 0 \ \quad \forall \ \, c=1,\ldots, C \ \textsuperscript{\ref{h_c}}
\end{equation}
becomes valid. It is important that these constraints must simultaneously hold for each dimension\textsuperscript{\ref{h_c}}. Note that 
\begin{equation}
w^c \geq \frac{w^c - f^c + 2P^c}{S^c} + 1 \quad \quad \forall \ \, c=1,\ldots, C \ \textsuperscript{\ref{h_c}}
\end{equation}
needs to be fulfilled in order for the size of the output maps not to become larger - this would add redundancy with many zero pads thus increasing computation. High level machine learning libraries are capable to automatically adjust (\ref{formula}) by adding zero-padding or cutting down $w^c$ and $h^c$ to make it fit \cite{cs231:optimizers}.\\
The calculations (\ref{firstmap_firstkernel_rowone_columnone})-(\ref{firstmap_firstkernel_rowone_columnlast})
define the convolutions when the filter slides column for column through the first row. Then the kernel slides column for column through the second row ((\ref{firstmap_firstkernel_rowtwo_columnone})-(\ref{firstmap_firstkernel_rowtwo_columnlast})).
This continues until the kernel convolutes through the last row ((\ref{firstmap_firstkernel_rowlast_columnone})-(\ref{firstmap_firstkernel_rowlast_columnlast})).
This procedure (\ref{firstmap_firstkernel_rowone_columnone})-(\ref{firstmap_firstkernel_rowlast_columnlast})
represents how a kernel (\ref{forallkernels}) convolutes entirely over an input map (\ref{forallinputmaps}) and produces an output map (\ref{foralloutputmaps}) $B_\delta^c$  (see figure (\ref{kernel}) for illustration). Then the kernel performs the entire convolution upon all next input maps (\ref{forallinputmaps}), and then the whole procedure is done all over again with the next kernel (\ref{forallkernels}). The result are $B^c_1, \ldots, B^c_{m^c \cdot \alpha(A^c)}$ output maps (\ref{foralloutputmaps}) each of which have size $\left(\frac{w^c - f^c + 2P^c}{S^c} + 1\right) \times \left(\frac{h^c - f^c + 2P^c}{S^c} + 1\right)$ and $B_\delta^c( \cdot, \cdot) \in \mathbb{R}$ for all $\delta = 1, \ldots, m^c \cdot \alpha(A^c)$ and $c=1, \ldots, C$.

\paragraph{ReLU} The Rectified Linear Unit Layer (ReLU) makes all values of the convoluted image non-negative. It is an activation function defined in (\ref{relu}) on page \pageref{relu}. ReLU is not really a separate layer, but more an operation performed on the previous convolutional layer. Commonly it will be performed right after the convolutional layer and therefore counts as part of the convolutional layer \cite{Goodfellow-et-al-2016}.

\begin{figure}[!htb]
\vskip 0.2in
\begin{center}
\centerline{\includegraphics[width=\columnwidth]{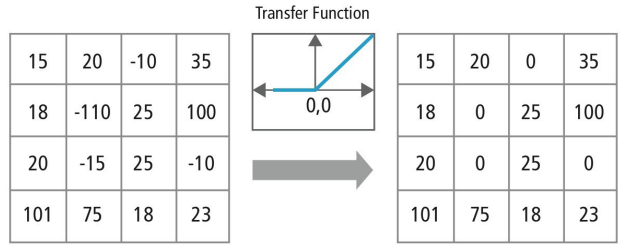}}
\caption{Illustration of ReLU in CNNs \cite{relu}}
\end{center}
\vskip -0.2in
\end{figure}

Let 
\begin{align}
B_1^c, \ldots, B_{ m^c \alpha(A^c)}^c \in \mathbb{R}^{\left({\frac{w^c - f^c + 2P^c}{S^c}+1}\right) \times \left({\frac{h^c - f^c + 2P^c}{S^c}+1}\right)}
\end{align} be the output maps of a convolutional layer $c$ and its corresponding hyperparameters, as defined in (\ref{firstmap_firstkernel_rowone_columnone}) - (\ref{firstmap_firstkernel_rowlast_columnlast}).\\
We will use the notation of (\ref{MILP_model2}) on page \pageref{MILP_model2} to model ReLU:

\begin{align}
&\overbrace{ 
\sum_{i=1}^{\frac{h^c - f^c + 2P^c}{S^c}+1} \sum_{j=1}^{{\frac{w^c - f^c + 2P^c}{S^c}+1}} B_\delta^c (i,j) \ = \ \hat{B}_\delta^c(i,j) - s^c_\delta(i,j) } \label{CNN_relu_activation} \\
& \hat{B}_\delta^c (i,j) \ , \ s_\delta^c(i,j) \ \geq \ 0 \notag \\
&z_\mu^c \in \{0, 1 \} \notag \\
& z_\mu^c (i,j) \, = \, 1 \ \to \ \hat{B}_\delta^c (i,j) \  \leq \ 0 \notag \\
&\underbrace{ z_\mu^c (i,j) \, = \, 0 \ \to \ s_\delta^c (i,j) \ \leq \ 0 \notag \qquad \qquad \qquad \qquad \qquad \
 } \\
&\forall \ \mu=1, \ldots, \left(\frac{h^c-f^c+2P^c}{S^c}+1\right)\cdot \left(\frac{w^c-f^c+2P^c}{S^c}+1\right) \\
&\forall \ \delta=1, \ldots, m^c \cdot \alpha(A^c), \quad \forall \ c=1, \ldots, C.
\end{align}

Concretely, we introduce an activation variable $z_\mu^c(i,j) \in \{0, 1 \}$ for every unit of map $B_\delta^c$, for every map $\delta =1, \ldots, m^c \cdot \alpha(A^c)$ and every convolutional layer $c=1, \ldots, C$. Similar to (\ref{linear_conditions}) on page \pageref{linear_conditions}, the activation value equals $1$ if the associated $B_\delta^c(i,j)$ is negative, thus turning the associated $B_\delta^c(i,j)$ into $0$. Else, if the associated $B_\delta^c(i,j)$ is positive, the activation value is $0$, thus legitimizing the associated $B_\delta^c(i,j)$ to be $\hat{B}_\delta^c(i,j)$.

\subsubsection{Pooling layer} 

\begin{figure}[!htb]
\vskip 0.2in
\begin{center}
\centerline{\includegraphics[width=\columnwidth]{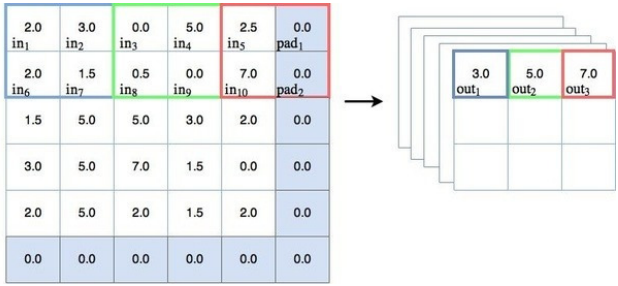}}
\caption{Illustration of max pooling \cite{maxpool}\label{max_pooling_pic}}
\end{center}
\vskip -0.2in
\end{figure}

The function of pooling layers in a CNN architecture, is to down-size the input maps in order to reduce the amount of parameters to avoid overfitting and reduce computation \cite{Goodfellow-et-al-2016}.\\
Pooling is similar to a kernel: it is an operation that takes pixel clusters of the input maps and combines it to one single pixel in the next layer, as can be seen in figure (\ref{max_pooling_pic}). Unlike convolutional kernels, pooling kernels do not have any component values.
There are several pooling functions, we will focus on max pooling as it is commonly used. Max pooling will take the maximum value of a kernel-sized pixel cluster and project it onto the output tensor. These maximum pixel values represent the most dominant and evident shapes in the original input map, only they are passed along and the rest of the pixel information is disregarded \cite{cs231:optimizers}.

\paragraph{Notation}

Let $1,\ldots, \mathcal{P}$ be the number of max pooling layers in the CNN. Each max pooling layer $p \in \{1,\ldots, \mathcal{P} \}$ consists of one max pooling kernel of which each has a predefined squared size $\mathfrak{f}^p \times \mathfrak{f}^p \in \mathbb{N}\times \mathbb{N}$. Note that $\mathfrak{f}^p$ is a hyperparameter and needs to be chosen for each layer $p$. Commonly $\mathfrak{f}^p \in \{2,3\} \quad \forall \ p=1, \ldots, \mathcal{P}$, or else too much information is lost \cite{cs231:optimizers}.\\
Suppose that $\mathcal{A}^p$ denotes the input of a pooling layer $p$ and $\mathcal{A}^p_1, \ldots, \mathcal{A}^p_{\alpha(\mathcal{A}^p)}$ are the input maps, where $\alpha(\mathcal{A}^p) \in \mathbb{N}$ is the number of input maps of $p$. Let $\mathcal{A}_\beta^p$ have size $\mathfrak{w}^p \times \mathfrak{h}^p$ for all $\beta=1, \ldots \alpha(\mathcal{A}^p)$ and $p \in \{1, \ldots \mathcal{P} \}$.\\
We can model a max pooling layer as follows:

\begin{align}
&\textbf{for} \quad \beta \ = \ 1, \ldots, \alpha(\mathcal{A}^p) \label{forall_inputmaps_pool}\\
&\overbrace{\mathcal{B}^p_\beta(1,1) \ = \ \max \left\lbrace \ \mathcal{A}_\beta^p(i,j) \quad | \quad \forall \ i,j=1, \ldots, \mathfrak{f}^p \right\rbrace  \qquad}\label{maxpool_firstrow_firstcolumn}\\
&\mathcal{B}^p_\beta(1,2) \ = \ \max \left\lbrace \ \mathcal{A}_\beta^p(i,j+\mathcal{S}^p) \quad | \quad \forall \ i,j=1, \ldots, \mathfrak{f}^p \right\rbrace \label{maxpool_firstrow_secondcolumn}\\
&\vdots \notag \\
&\mathcal{B}^p_\beta\left(1,\frac{\mathfrak{w}^p -\mathfrak{f}^p}{\mathcal{S}^p}+1\right) \ = \notag \\
&\max \left\lbrace \ \mathcal{A}_\beta^p(i,\mathfrak{w}^p - \mathfrak{f}^p +j) \quad | \quad \forall \ i,j=1, \ldots, \mathfrak{f}^p \right\rbrace \label{maxpool_firstrow_lastcolumn} \\
&\rule{80mm}{.01pt} \notag \\
&\mathcal{B}^p_\beta(2,1) \ = \ \max \left\lbrace \ \mathcal{A}_\beta^p(i+\mathcal{S}^p,j) \quad | \quad \forall \ i,j=1, \ldots, \mathfrak{f}^p \right\rbrace \label{maxpool_secondrow_firstcolumn}\\
&\vdots \notag \\
&\mathcal{B}^p_\beta\left(2,\frac{\mathfrak{w}^p -\mathfrak{f}^p}{\mathcal{S}^p}+1\right) \ = \notag \\
&\max \left\lbrace \ \mathcal{A}_\beta^p(i + \mathcal{S}^p,\mathfrak{w}^p - \mathfrak{f}^p +j) \quad | \quad \forall \ i,j=1, \ldots, \mathfrak{f}^p \right\rbrace \label{maxpool_secondrow_lastcolumn} \\
&\rule{80mm}{.01pt} \notag \\
&\vdots \qquad \qquad \qquad \qquad \qquad  \vdots \qquad \qquad \qquad \qquad  \qquad \qquad \vdots \notag \\
&\rule{80mm}{.01pt} \notag \\
&\mathcal{B}^p_\beta\left( \frac{\mathfrak{h}^p - \mathfrak{f}^p}{\mathcal{S}^p}+1,1\right) \ = \notag \\
&\max \left\lbrace \ \mathcal{A}_\beta^p(\mathfrak{h}^p - \mathfrak{f}^p+i,j) \quad | \quad \forall \ i,j=1, \ldots, \mathfrak{f}^p \right\rbrace \label{maxpool_lastrow_firstcolumn} \\
&\vdots \notag \\
&\mathcal{B}^p_\beta\left( \frac{\mathfrak{h}^p - \mathfrak{f}^p}{\mathcal{S}^p}+1,\frac{\mathfrak{w}^p - \mathfrak{f}^p}{\mathcal{S}^p}+1\right) \ = \notag \\
&\underbrace{\max \left\lbrace \ \mathcal{A}_\beta^p(\mathfrak{h}^p - \mathfrak{f}^p+i,\mathfrak{w}^p - \mathfrak{f}^p+j) \quad | \quad \forall \ i,j=1, \ldots, \mathfrak{f}^p \right\rbrace }\label{maxpool_lastrow_lastcolumn}  \\
& \qquad \qquad \qquad \qquad \forall \ p \ \in \{1, \ldots, \mathcal{P} \}, \notag
\end{align}
where $\mathcal{S}^p \in \mathbb{N}$ defines the specified stride length of each max pooling layer $p \in \{ 1, \ldots, \mathcal{P} \}$.  For each stride, $\mathcal{S}^p \leq \mathfrak{f}^p$ must be valid for all $p=1, \ldots, \mathcal{P}$, otherwise input pixels are skipped \cite{cs231:optimizers}. The stride needs to be chosen in a way that\footnote{same for height $\mathfrak{h}^p$ respectively.\label{mathfrak_h^p}}
\begin{align}
\frac{\mathfrak{w}^p- \mathfrak{f}^p}{\mathcal{S}^p} + 1 \quad \forall \ p=1, \ldots, \mathcal{P} \ \textsuperscript{\ref{mathfrak_h^p}}
\end{align} 
is integer, in other words such that
\begin{align}
\mathfrak{w}^p - \mathfrak{f}^p \mod  \ \mathcal{S}^p \ = \ 0 \quad \forall \ p=1, \ldots, \mathcal{P}.\textsuperscript{\ref{mathfrak_h^p}}
\end{align}
Given a $\beta \in 1, \ldots, \alpha(\mathcal{A}^p)$, we are given the input map $\mathcal{A}_\beta^p$ (\ref{forall_inputmaps_pool}), on which we apply the max pooling kernel at the first row through all columns (\ref{maxpool_firstrow_firstcolumn})-(\ref{maxpool_firstrow_lastcolumn}) and storing the max value of each $\mathfrak{f}^p \times \mathfrak{f}^p$-pixel cluster of  $\mathcal{A}^p_\beta$ as a single value $\mathcal{B}_\beta^p(1, \cdot)$. Then max pooling is performed at the second row through all  columns (\ref{maxpool_secondrow_firstcolumn})-(\ref{maxpool_secondrow_lastcolumn}).  and the max values are stored in $\mathcal{B}_\beta^p(2, \cdot)$. This continues for all rows, until finally max pooling is done on the last row through all columns (\ref{maxpool_lastrow_firstcolumn})-(\ref{maxpool_lastrow_lastcolumn})
with the corresponding max values stored in $\mathcal{B}_\beta^p(\mathfrak{f}^p, \cdot)$. This whole procedure is done for all input maps (\ref{forall_inputmaps_pool}).\\
The result of a max pooling layer $p \in \{1, \ldots, \mathcal{P} \}$ are $\alpha(\mathcal{A}^p)$ output maps $\mathcal{B}_1^p, \ldots, \mathcal{B}_{\alpha(\mathcal{A}^p)}^p$, each of which have size $\left( \frac{\mathfrak{w}^p - \mathfrak{f}^p}{\mathcal{S}^p}+1 \right) \times \left( \frac{\mathfrak{h}^p - \mathfrak{f}^p}{\mathcal{S}^p}+1 \right)$ and $\mathcal{B}^p_\beta(\cdot, \cdot) \in \mathbb{R}$ for all $\beta=1, \ldots, \alpha(\mathcal{A}^p)$.\\
\\
Modelling max pooling in a way that it can be used by modern MILP solvers, requires the use of binary activation variables e.g. given the first max pooling operation (\ref{maxpool_firstrow_firstcolumn}), we can transform it into:
\begin{align}
&\overbrace{\sum_{\mu =1}^{\mathfrak{f}^{2\cdot p}} z_\mu ^p \ = \ 1 \qquad \qquad \qquad\qquad} \label{sum_of_all_actiation_variables}\\
& \mathcal{B}_\beta^p(1,1) \ \geq \ \ \mathcal{A}_\beta^p(i,j) \label{constraint1} \\
& z_\mu^p = 1 \ \to \ \mathcal{B}_\beta^p(1,1) \ \leq \ \mathcal{A}_\beta^p(i,j) \label{constraint2} \\
& \underbrace{ z_\mu^p \in \{0, 1\} \qquad \qquad \qquad \qquad \quad} \label{constraint3}\\
& \forall \ i,j = 1, \ldots, \mathfrak{f}^p \label{constraint4}\\
& \forall \ \beta = 1, \ldots, \alpha(\mathcal{A}^p), \quad \forall \
p \in \{ 1, \ldots, \mathcal{P} \} \label{constraint5}.
\end{align}
For this first max pooling kernel, we assign $\mathfrak{f}^{2\cdot p}$ times activation variables $z_\mu^p$. They correspond to the first pixel cluster of the input map $\mathcal{A}_\beta^p$ - only one of them will be the maximum value (\ref{sum_of_all_actiation_variables}), thus having activation value $1$, while the rest have $0$. The maximum value will satisfy (\ref{constraint1}) for all $i,j$ (\ref{constraint4}). Specifically $\mathcal{B}_\beta^p(1,1)$ is equals to the maxium, as to which the activation variable triggers (\ref{constraint2}) - the remaining activation variables equal $0$ (\ref{constraint3}). This works for every input map (\ref{constraint5}) and every max pooling layer (\ref{constraint5}).

Similarly we can construct such MILP constraints for every max pooling operation (\ref{maxpool_firstrow_firstcolumn})-(\ref{maxpool_lastrow_lastcolumn}), which we will write more generically in section (\ref{CNN_MILP_formulation}) on page \pageref{CNN_MILP_formulation}.

\subsubsection{Fully-Connected layer} All the pixel representations of the last layer (either convolutional or pooling) are reshaped into one long unit layer - the so called flattend layer \cite{wiki_convnets}. Then each unit of the flattend layer connects entirely to the fully-connected layer. Unlike a convolutional layer, the units in the fully-connected layer do not share weights (\cite{cs231:optimizers}).

\paragraph{Notation}
Let $\mathfrak{A}_1, \ldots, \mathfrak{A}_{\alpha(\mathfrak{A})}$ be the output of the last convolutional or pooling layer of the CNN. Suppose $\eta \times \omega$ is the size of each $\mathfrak{A}_{\beta}$ with $\beta=1, \ldots, \alpha(\mathfrak{A})$. Define
\begin{align}
\pi\left(\beta \cdot ( \omega \cdot (i-1) + j)\right) \ = \ \mathfrak{A}_\beta(i,j)
\end{align}
for all $i=1, \ldots, \eta$, $j=1, \ldots, \omega$ and $\beta =1, \ldots, \alpha(\mathfrak{A})$. This gives a flattend layer $\pi$ of size $\mathbb{R}^{\left(\alpha(\mathfrak{A})\cdot\eta\cdot\omega\right) \times 1}$ and represents merely a reshape of the pixel maps $\mathfrak{A}_1, \ldots, \mathfrak{A}_{\alpha(\mathfrak{A})}$ into a single long unit layer.\\
Let the next layer be the fully-connected layer $\phi$ with size $\mathbb{R}^{n_{\phi} \times 1}$, where $n_{\phi} \in \mathbb{N}$ is the number of units of the fully-connected layer. Since we are one layer behind the output class layer, it is advisable to choose $n_{\phi} < \alpha(\mathfrak{A})\cdot \eta \cdot \omega$, we want to decrease the units to get closer to the number of units in our output class layer. The flattend layer is fully-connected with the fully-connected layer i.e. there are trainable weight parameters on each edge connecting each flattend layer unit with every fully-connected layer unit. Set $W^{n_\phi \times \left(\alpha(\mathfrak{A})\cdot\eta \cdot \omega\right)}_\pi \in \mathbb{R}^{n_\phi \times \left(\alpha(\mathfrak{A})\cdot\eta \cdot \omega\right)}$ to associate the weights on all edges between $\pi$ and $\phi$, concretely $w_\pi(i,j) \in \mathbb{R}$ is the weight between unit $i$ of the flattend layer and unit $j$ of the fully-connected layer. The value of a fixed unit $i$ of $\phi$ is calculated by
\begin{align}
\phi(i) \ = \ \text{ReLU} \left( \sum_{k=1}^{\alpha(\mathfrak{A}) \cdot \eta \cdot \omega} \pi(k) \cdot w_\pi (k,i) \right) \quad \forall \ i=1, \ldots, n_\phi,
\end{align}
we choose ReLU to be the activation function for $\phi$.

\subsubsection{Output layer}

The output class layer $\psi \in (0,1]^{n_\psi \times 1}$ is the final layer of the CNN and corresponds to the classification of the input image, where $n_\psi$ is the number of classes of the CNN. This layer is fully-connected to $\phi$, i.e. we associate all weights between layer $\phi$ and $\psi$ as weight matrix $W_\phi^{n_\psi \times n_\phi} \in \mathbb{R}^{n_\psi \times n_\phi}$. The value of a fixed unit $i$ of $\psi$ is calculated by
\begin{align}
\psi(i) \ = \ \text{softmax} \left( \sum_{k=1}^{n_\phi} \phi(k) \cdot w_\phi (k,i) \right) \quad \forall \ i=1, \ldots, n_\psi,
\end{align}
where softmax \cite{wiki_softmax} is defined by
\begin{align}
&\text{softmax} \left( \sum_{k=1}^{n_\phi} \phi(k) \cdot w_\phi (k,i) \right) \ = \\
&\frac{\exp \left( \sum_{k=1}^{n_\phi} \phi(k) \cdot w_\phi (k,i) \right)}{\sum_{i=1}^{n_\psi} \exp \left( \sum_{k=1}^{n_\phi} \phi(k) \cdot w_\phi (k,i) \right)}. \label{softmax}
\end{align}
Each entry is a value between $(0,1]$ and all entries add up to $1$. Applying softmax as activation function to $\psi$ allows the entries to be activation values, thus allows to interpret the activation of the output class layer as probability predictions of each class.\\
This part of the CNN is similar to a standard neural network and every methode and technique, which applies to standard vanilla neural nets, can also be applied to this part e.g. softmax as activation, dropout measures etc \cite{cs231:optimizers}. Each output neuron will give a prediction in form of a probability as to whether this object class is recognized in the image. The highest probability (closest to 1) will be the total prediction for the image \cite{wiki_convnets}.\\

\subsection{CNN training}

\begin{figure}[!htb]
\vskip 0.2in
\begin{center}
\centerline{\includegraphics[width=\columnwidth]{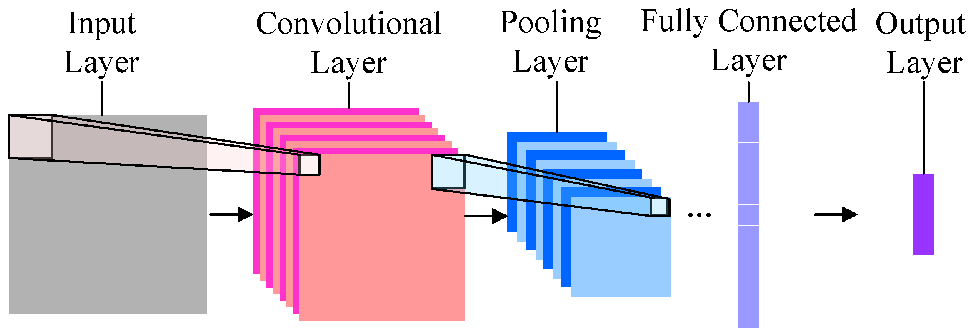}}
\caption{\cite{sequence}}\label{basic_CNN_structure}
\end{center}
\vskip -0.2in
\end{figure}

Choosing a CNNs layers and hyperparameters remains an uncertainty, commonly however there are rules of thumb to the hyperparameters \cite{cs231:optimizers}. A basic sequence of CNN layers can be seen in figure (\ref{basic_CNN_structure}).

\begin{figure}[ht]
\vskip 0.2in
\begin{center}
\centerline{\includegraphics[width=\columnwidth]{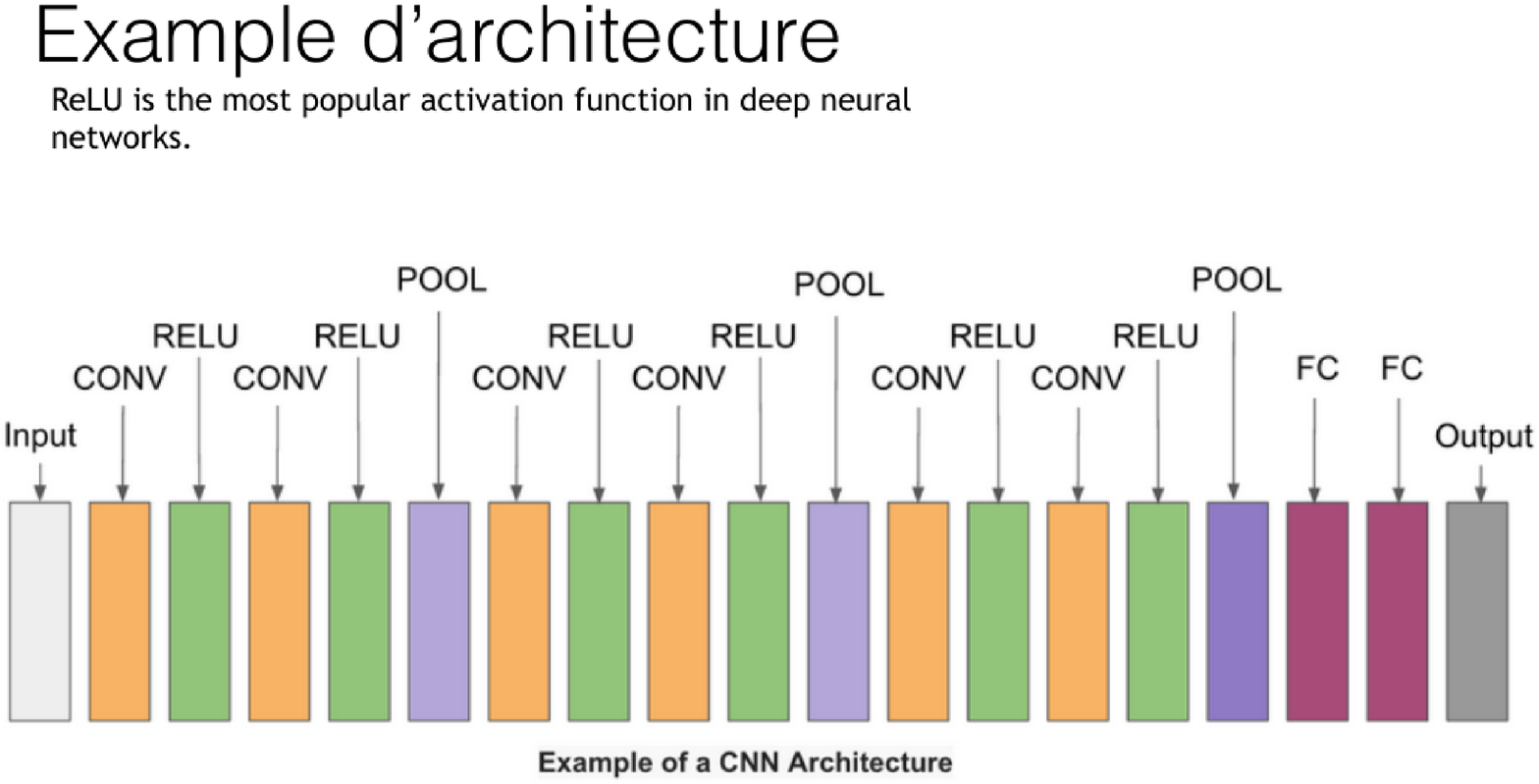}}
\caption{\cite{chevalyre}\label{chevalyre}}
\end{center}
\vskip -0.2in
\end{figure}

A CNN can hold several convolutional and max pooling layers, which alternate, as in figure (\ref{chevalyre}) This way, a CNN can be build to a much deeper architecture and hopes are that performance will be more accurate - not necessarily though due to overfitting and many more options to tweak \citep{performance_tweaks}.

Similarily to DNNs, the values of the convolutional filters as well as the weight matrices of the fully-connected layer are trainable paramaters for CNNs. We use backpropagation and gradient descent to improve each of the filters weights \citep{uniform_distribut} (\ref{DNN_training}). Cross-entropy as loss function is commonly used \cite{cs231:optimizers}.

\subsection{MILP formulation for a CNN}
\label{CNN_MILP_formulation}
Analogeously to designing a $0$-$1$ MILP formulation for DNNs (\ref{MILP_model1})-(\ref{MILP_model3}), we can formulate a $0$-$1$ MILP representation for CNNs.\\
We set several assumptions:
\begin{itemize}
\item The CNN consists of $C$ layers and each $c\in \{ 1, \ldots, C \}$ is is a \textbf{block}: a convolutional layer (including ReLU) is followed rigidly by a max pooling layer i.e. the input of $c$ is the output of the max pooling layer of block $c-1$ and the output of $c$ is the input of the convolutional layer of block $c+1$. This allows the MILP to be written in an iterative structure.\\
The consequence is we have hyperparameters $\mathcal{S}^c, \mathfrak{f}^c, \ldots$ where superscript $c$ points that this pooling layer is part of block $c$. Note that $\mathcal{P} = C$ for this MILP. The MILP is customizable to allow a more dynamic use of alternating layers.
\item The flattend layer $\pi$, the fully-connected layer $\phi$ and the output class layer $\psi$ are not part of a block and are defined separately. The flattend layer follows after the last block $C$.
\item For the sake of simplicity, bias units are omitted.
\item For the sake of simplicity, we introduce substitutions (\ref{73})-(\ref{76}).
\item It is problematic, that the use of the non-linear softmax activation function cannot be used in a MILP model. Due to the unchanged derivative the linear approximation does not work either. This also rules out the sigmoid function as a possible alternative. This is why we will use ReLU in the output class layer as activation.
\item Every convolutional kernel value $k_\gamma^c(i,j)$ for all $i,j=1,\ldots, f^c$, $\gamma=1, \ldots, m^c$ and all blocks $c$ are \textit{given} values.
\item The weight parameters $w_\pi(\mathfrak{k}, i_\phi)$ and $w_\psi(\mathfrak{k}, i_\psi)$ for all $\mathfrak{k}=1, \ldots, m^C \cdot \alpha(A^C) \cdot \tilde{\tilde{h}}^{C} \cdot \tilde{\tilde{w}}^{C}$, $i_\phi = 1, \ldots, n_\phi$ and $i_\psi = 1, \ldots, n_\psi$ for the fully-connected layer and the output class layer respectively are \textit{given} values.
\item Hyperparameters to set:
\begin{itemize}
\item The number of blocks $C$ consisting of a convolutional layer, ReLU activation and max pooling.
\item The size $h^1 \times w^1$ of the input images.
\item The number of input channels $\alpha(A^1)$.
\item The number of convolutional kernels $m^c$ for each block $c$.
\item The size of the convolutional kernels $f^c$ for each block $c$.
\item The size of the convolutional stride $S^c$ for each block $c$.
\item The number of convolutional zero-padding $P^c$ for each block $c$.
\item The size of max pooling kernel $\mathfrak{f}^c$ for each block $c$.
\item The size of the max pooling stride $\mathcal{S}^c$ for each block $c$.
\item The number of units $n_\phi$ in the fully-connected layer .
\item The number of classes $n_\psi$ in the output class layer.
\end{itemize}
\end{itemize}

\begin{align}
& \text{substitutions:} \notag \\
& \tilde{h}^c \ := \  \left( \frac{h^c - f^c +2P^c}{S^c}+1 \right) \label{73}\\
& \tilde{w}^c \ := \  \left( \frac{w^c - f^c +2P^c}{S^c}+1 \right) \label{74}\\
& \tilde{\tilde{h}}^c \ := \  \left( \frac{\tilde{h}^c - \mathfrak{f}^c }{\mathcal{S}^c}+1 \right) \label{75}\\
& \tilde{\tilde{w}}^c \ := \  \left( \frac{\tilde{w}^c - \mathfrak{f}^c}{\mathcal{S}^c}+1 \right)  \label{76}
\end{align}

\begin{align}
&\min \quad \Bigg( \sum_{c=1}^C \ \Bigg( \quad \sum_{\beta=1}^{\alpha(A^c)} \sum_{i=\lambda}^{h^c} \sum_{\xi=1}^{w^c} c_\beta^c(\lambda,\xi) \cdot A_\beta^c(\lambda,\xi) \ + \label{obj_fct_conv_77} \\
& \sum_{\delta=1}^{m^c \cdot \alpha(A^c)} \sum_{\tilde{\lambda}=1}^{\tilde{h}^c} \sum_{\tilde{\xi} = 1}^{\tilde{w}^c} g_\delta^c(\tilde{\lambda}, \tilde{\xi}) \cdot B_\delta^c(\tilde{\lambda}, \tilde{\xi}) \ + \label{obj_fct_conv_78} \\
& \sum_{\delta=1}^{m^c \cdot \alpha(A^c)} \sum_{\tilde{\lambda}=1}^{\tilde{h}^c} \sum_{\tilde{\xi} = 1}^{\tilde{w}^c} l_\delta^c(\tilde{\lambda},\tilde{\xi}) \cdot \hat{B}_\delta^c(\tilde{\lambda}, \tilde{\xi}) \ + \label{obj_fct_conv_79} \\
& \sum_{\delta = 1}^{m^c \cdot \alpha(A^c)} \sum_{\tilde{\lambda}=1}^{\tilde{h}^c} \sum_{\tilde{\xi}=1}^{\tilde{w}^c} n_\delta^c(\tilde{\lambda},\tilde{\xi}) \cdot  z_\delta^c(\tilde{\lambda},\tilde{\xi}) \ + \label{obj_fct_conv_80} \\
&\sum_{\delta = 1}^{m^c \cdot \alpha(A^c)} \sum_{\triangle = 1}^{\tilde{\tilde{h}}^c \cdot \tilde{\tilde{w}}^c} \sum_{\tilde{\tilde{\lambda}}=1}^{\tilde{\tilde{h}}^c} \sum_{\tilde{\tilde{\xi}}=1}^{\tilde{\tilde{w}}^c} o_{\triangle, \delta}^c(\tilde{\tilde{\lambda}}, \tilde{\tilde{\xi}}) \cdot \zeta_{\triangle, \delta}^c(\tilde{\tilde{\lambda}},\tilde{\tilde{\xi}}) \ \Bigg) \label{obj_fct_conv_81}\\
&+ \ \sum_{i_\phi = 1}^{n_\phi} \left( \tilde{c}(i_\phi) \cdot \phi(i_\phi) \ + \ q(i_\phi) \cdot \tilde{\zeta}(i_\phi) \right) \ + \ \label{obj_fct_conv_82}\\
&\sum_{i_\psi = 1}^{n_\psi}  \tilde{\tilde{c}}(i_\psi) \cdot \psi(i_\psi) \Bigg) \label{obj_fct_conv_83} \\
\notag \\
\notag \\
&B_\delta^c(\tilde{\lambda},\tilde{\xi}) \ = \ \notag \\
&\sum_{i=1}^{f^c} \, \sum_{j=1}^{f^c} \ A_\beta^c\left((i+S^c\cdot(\tilde{\lambda}-1),j+S^c\cdot(\tilde{\xi} -1)) \cdot k_\gamma^c(i,j)\right) \label{const_conv_84}\\
\notag \\
\notag \\
&\sum_{\tilde{\lambda}=1}^{\tilde{h}^c} \sum_{\tilde{\xi}=1}^{\tilde{w}^c} B_\delta^c (\tilde{\lambda},\tilde{\xi}) \ = \ \hat{B}_\delta^c(\tilde{\lambda},\tilde{\xi}) - s^c_\delta(\tilde{\lambda},\tilde{\xi})  \label{const_relu_85} \\
& \hat{B}_\delta^c (\tilde{\lambda},\tilde{\xi}) \ , \ s_\delta^c(\tilde{\lambda},\tilde{\xi}) \ \geq \ 0 \label{const_relu_86} \\
&z_\delta^c(\tilde{\lambda},\tilde{\xi}) \in \{0, 1 \} \label{const_relu_87}\\
& z_\delta^c (\tilde{\lambda},\tilde{\xi}) \, = \, 1 \ \to \ \hat{B}_\delta^c (\tilde{\lambda},\tilde{\xi}) \  \leq \ 0 \label{const_relu_88} \\
& z_\delta^c (\tilde{\lambda},\tilde{\xi}) \ = \ 0 \ \to \ s_\delta^c (\tilde{\lambda},\tilde{\xi}) \ \leq \ 0 \label{const_relu_89} \\
\notag \\
\notag \\
&\sum_{\mathfrak{i} =1}^{\mathfrak{f}^c} \sum_{\mathfrak{j}=1}^{\mathfrak{f}^c} \zeta_{\triangle, \delta}^c(\mathfrak{i},\mathfrak{j}) \ = \ 1 \label{const_pool_90} \\
& A_\delta^{c+1}(\tilde{\tilde{\lambda}}, \tilde{\tilde{\xi}}) \ \geq \ \ \hat{B}_\delta^c(\mathfrak{i}+\mathcal{S}^c\cdot(\tilde{\tilde{\lambda}} - 1),\mathfrak{j}+\mathcal{S}^c \cdot( \tilde{\tilde{\xi}} - 1)) \label{const_pool_91}  \\
&  \zeta_{\triangle, \delta}^c(\mathfrak{i},\mathfrak{j}) \ = \ 1 \to \notag \\
& A_\delta^{c+1}(\tilde{\tilde{\lambda}}, \tilde{\tilde{\xi}}) \ \leq \ \hat{B}_\delta^c(\mathfrak{i}+\mathcal{S}^c\cdot(\tilde{\tilde{\lambda}} - 1),\mathfrak{j}+\mathcal{S}^c \cdot( \tilde{\tilde{\xi}} - 1)) \label{const_pool_92}  \\
& \zeta_{\triangle, \delta}^c(\mathfrak{i},\mathfrak{j}) \in \{0, 1\} \label{const_pool_93}  \\
\notag \\
&\pi \left(\tilde{\delta}\cdot( \tilde{w}^{C} \cdot (\tilde{\tilde{\lambda}}-1) + \tilde{\tilde{\xi}} )\right) \ = \ A_{\tilde{\delta}}^{C+1}(\tilde{\tilde{\lambda}}, \tilde{\tilde{\xi}}) \label{const_flatten_94} \\
\notag \\
&\sum_{\mathfrak{k}=1}^{m^{C}\cdot \alpha(A^{C}) \cdot \tilde{\tilde{h}}^{C} \cdot \tilde{\tilde{w}}^{C}} \pi(\mathfrak{k}) \cdot w_\pi (\mathfrak{k}, i_\phi) \ = \ \phi(i_\phi) - \tilde{s}(i_\phi) \label{const_fullyconnect_relu_95} \\
& \phi(i_\phi), \, \tilde{s}(i_\phi) \ \geq \ 0 \label{const_fullyconnect_relu_96}\\
& \tilde{\zeta}(i_\phi) \in \{ 0,1 \} \label{const_fullyconnect_relu_97}\\
& \tilde{\zeta}(i_\phi) = 1  \ \to \ \phi(i_\phi) \leq 0 \label{const_fullyconnect_relu_98} \\
& \tilde{\zeta}(i_\phi) = 0  \ \to \ \tilde{s}(i_\phi) \leq 0 \label{const_fullyconnect_relu_99}\\
\notag \\
&\sum_{\mathfrak{k}=1}^{m^{C}\cdot \alpha(A^{C}) \cdot \tilde{\tilde{h}}^{C} \cdot \tilde{\tilde{w}}^{C}} \phi(\mathfrak{k}) \cdot w_\phi (\mathfrak{k}, i_\psi) \ = \ \psi(i_\psi) - \tilde{\tilde{s}}(i_\psi) \label{const_output_100} \\
& \psi(i_\psi), \, \tilde{\tilde{s}}(i_\psi) \ \geq \ 0 \label{const_output_101} \\
& \tilde{\tilde{\zeta}}(i_\psi) \in \{ 0,1 \} \label{const_output_102}\\
& \tilde{\tilde{\zeta}}(i_\psi) = 1  \ \to \ \psi(i_\psi) \leq 0 \label{const_output_103}\\
& \tilde{\tilde{\zeta}}(i_\psi) = 0  \ \to \ \tilde{\tilde{s}}(i_\psi) \leq 0  \label{const_output_104}\\
\notag \\
&lb(A_\beta^c)(\lambda, \xi) \ \leq \ A_\beta^c(\lambda,\xi) \ \leq \ ub(A_\beta^c)(\lambda, \xi) \label{const_bounds_105} \\
&lb(B_\delta^c)(\tilde{\lambda}, \tilde{\xi}) \ \leq \ B_\delta^c(\tilde{\lambda}, \tilde{\xi}) \ \leq \ ub(B_\delta^c)(\tilde{\lambda}, \tilde{\xi}) \label{const_bounds_106} \\
&lb(\hat{B}_\delta^c)(\tilde{\lambda}, \tilde{\xi}) \ \leq \ \hat{B}_\delta^c(\tilde{\lambda}, \tilde{\xi}) \ \leq \ ub(\hat{B}_\delta^c)(\tilde{\lambda}, \tilde{\xi}) \label{const_bounds_107} \\ 
&lb(\pi)(\mathfrak{k}) \ \leq \ \pi(\mathfrak{k}) \ \leq \ ub(\pi)(\mathfrak{k}) \label{const_bounds_108}\\
&lb(\psi)(\mathfrak{k}) \ \leq \psi(\mathfrak{k}) \ \leq \ ub(\psi)(\mathfrak{k}) \label{const_bounds_109} \\
&lb(s_\delta^c)(\tilde{\lambda}, \tilde{\xi}) \ \leq \ s_\delta^c(\tilde{\lambda}, \tilde{\xi}) \ \leq \ ub(s_\delta^c)(\tilde{\lambda}, \tilde{\xi}) \label{const_bounds_110}\\
&lb(\tilde{s})(i_\phi) \ \leq \ \tilde{s}(i_\phi) \ \leq \ ub(\tilde{s})(i_\phi) \label{const_bounds_111}\\
&lb(\tilde{\tilde{s}})(i_\phi) \ \leq \ \tilde{\tilde{s}}(i_\phi) \ \leq \ ub(\tilde{\tilde{s}})(i_\phi) \label{const_bounds_112} \\
\notag \\
& \forall \ c \in \{ 1, \ldots, \mathcal{C} \} \label{forall:c_107} \\
&\forall \ \beta \ = \ 1, \ldots, \alpha(A^c) \label{forall:beta_108} \\
&\forall \ \lambda = 1,  \ldots, h^c \label{forall:lamda_109}\\
&\forall \ \xi = 1, \ldots, w^c \label{forall:xi_110}\\
&\forall \ i,j \ = \ 1,\ldots, f^c \label{forall:ij_111}\\
& \forall \ \delta = 1, \ldots, m^c \cdot \alpha(A^c) \label{forall:delta_112} \\
&\forall \ \tilde{\lambda}=1, \ldots, \tilde{h}^c \label{forall:lamdatilde_113} \\
&\forall \ \tilde{\xi}=1, \ldots, \tilde{w}^c \label{forall:xitilde_114} \\
&\forall \ \triangle =1, \ldots, \tilde{\tilde{h}}^c \cdot \tilde{\tilde{w}}^c \label{forall:triangle_115} \\
& \forall \ \tilde{\tilde{\lambda}}=1, \ldots, \tilde{\tilde{h}}^c \label{forall:lambdatildetilde_116}\\
& \forall \ \tilde{\tilde{\xi}}=1, \ldots, \tilde{\tilde{w}}^c \label{forall:xitildetilde:117}\\
&\forall \ \gamma \ = \ 1, \ldots, m^c \label{forall:gamma:116}\\
& \forall \ \mathfrak{i},\mathfrak{j} = 1, \ldots, \mathfrak{f}^c \label{forall:ij_117}\\
& \forall \ i_{\phi} = 1, \ldots, n_\phi \label{forall:iphi_118} \\
& \forall \ i_{\psi} = 1, \ldots, n_\psi \label{forall:ipsi_119}
\end{align}

This MILP formulation is feasible, since for any \textit{fixed} input $A_\beta^1$ i.e. $lb(A_\beta^1)(\lambda, \xi) = ub(A_\beta^1)(\lambda, \xi) \quad \forall \ \beta=1 \ldots \alpha(A^1), \ \forall \ \lambda=1, \ldots, h^1, \ \forall \ \xi = 1, \ldots, w^1$, every other unit in the system is uniquely defined by (\ref{const_conv_84})-(\ref{const_output_104}).

\subsubsection{Explanation}

This MILP model is designed to be a minimization problem (\ref{obj_fct_conv_77})-{\ref{obj_fct_conv_83}), in which the value of each unit in every layer is minimized.\\
Explanation of (\ref{obj_fct_conv_77}): as mentioned, this MILP is a minimization problem. The variables are \textit{mixed} i.e. they are real or binary integers. The objective function and the constraints must be linear. All variables of blocks $c=1, \ldots, C$ must be minimized, as well as all pixel values of each map, $A_\beta^c(\lambda,\xi)$, of all input maps $\beta$. The CNN's input maps are marked as $A_\beta^1(\lambda,\xi)$, the size is $h^1 \times w^1$. The cost parameters $c_\beta^c(\lambda,\xi) \in \mathbb{R}$ can be set as $1$ for all $c$, $\beta$, $\lambda$ and $\xi$ (\ref{forall:c_107})-(\ref{forall:xi_110}).\\
Explanation of (\ref{obj_fct_conv_78}): the units in $B_\delta(\tilde{\lambda}, \tilde{\xi})$ represent the convoluted units. The number of maps make up the number of input maps times the number of kernels used. The size of these maps have changed depending on the kernel size, the stride and the padding. The cost parameters $g_\delta^c(\tilde{\lambda},\tilde{\xi}) \in \mathbb{R}$ can be set as $1$ for all $c$, $\delta$, $\tilde{\lambda}$ and $\tilde{\xi}$ (\ref{forall:delta_112})-(\ref{forall:xitilde_114}).\\
Explanation of (\ref{obj_fct_conv_79}): the convoluted maps are fed through the ReLU function, which gives us the unit maps $\hat{B}_\delta^c(\tilde{\lambda}, \tilde{\xi})$, the size is unchanged. The cost parameters $l_\delta^c(\tilde{\lambda}, \tilde{\xi}) \in \mathbb{R}$ can be set as $1$ for all $c$, $\delta$, $\tilde{\lambda}$ and $\tilde{\xi}$ (\ref{forall:delta_112})-(\ref{forall:xitilde_114}).\\
Explanation of (\ref{obj_fct_conv_80}): the binary variable $z_\delta^c(\tilde{\lambda}, \tilde{\xi})$ corresponds as an activation variable used for the ReLU function. We penalize the occurrence of $z_\delta^c(\tilde{\lambda}, \tilde{\xi})=1$, we may initialize the cost parameters $n_\delta^c(\tilde{\lambda}, \tilde{\xi})$ in a way in which how much we want to penalize this occurrence, possibly between $[0,1]$.\\
Explanation of (\ref{obj_fct_conv_81}): this minimizes the binary activation variable $\zeta_{\triangle, \delta}^c(\tilde{\tilde{\lambda}},\tilde{\tilde{\xi}}) $ used for max pooling. For each pooling kernel $\triangle$ (\ref{forall:triangle_115}) in an input map $\delta$, we need to associate each entry $(\tilde{\tilde{\lambda}}, \tilde{\tilde{\xi}})$ (\ref{forall:lambdatildetilde_116})-(\ref{forall:xitildetilde:117}) with this $0$-$1$ activation variable. We may set the cost parameters e.g. $o_{\triangle, \delta}^c(\tilde{\tilde{\lambda}},\tilde{\tilde{\xi}}) \in [0,1]$.\\
Explanation of (\ref{obj_fct_conv_82}): the units in $\phi(i_\phi)$ represent the fully-connected layer $\phi$. We can choose the cost parameters $q(i_\phi)$ to be $1$. Furthermore $\tilde{\zeta}(i_\phi)$ stands for the activation variable used for ReLU in the fully-connected layer. Similar to $z$ and $\zeta$, we can choose the cost parameters $q(i_\phi) \in [0,1]$ depending on how much we want to penalize the activation.\\
Explanation of (\ref{const_conv_84}): the input maps to block $c$ are convoluted. The unit values are multiplied by a kernel value and all kernel values are summed up. This convolution depends on the kernel size $f^c$ (\ref{forall:ij_111}), the stride and padding measures. A kernel moves along the an input map horizontally, then vertically, and then performs this on all input maps $\beta$, before the next kernel (\ref{forall:gamma:116}) repeats this procedure. As a result we have $B_\delta^c(\tilde{\lambda}, \tilde{\xi})$ with changed size. \\
Explanation of (\ref{const_relu_85}): by introducing variable $s_\delta^c(\tilde{\lambda}, \tilde{\xi})$, we can tackle a non-linear activation function to fit into a MILP. For this we introduce an activation variable $z_\delta^c(\tilde{\lambda}, \tilde{\xi})$. The purpose of (\ref{const_relu_86})-(\ref{const_relu_89}) matches the thoughts of (\ref{z1})-(\ref{z01}).\\
Explanation of (\ref{const_pool_90}): we define an activation variable $\zeta_{\triangle, \delta}^c(\mathfrak{i}, \mathfrak{j})$ (\ref{const_pool_93}) for each input map $\delta$, we have $\triangle$ pooling kernels and each kernel has $\mathfrak{f}^c \times \mathfrak{f}^c$ (\ref{forall:ij_117}) components. We need to step through each component and guarantee that each kernel only activates one $\zeta$ (\ref{const_pool_90}). This activation identifies the max value of the pixel cluster in maps $\hat{B}_\delta^c$ (\ref{const_pool_91})-(\ref{const_pool_92}).\\
The constraints (\ref{const_conv_84})-(\ref{const_pool_93}) characterize the convolution-, the ReLU and the max pooling layers in a block $c=1, \ldots, C$. The last block $C$ will give us maps $A_\delta^{C+1}(\tilde{\tilde{\lambda}}, \tilde{\tilde{\xi}})$ for all $\tilde{\tilde{\lambda}}$ (\ref{forall:lambdatildetilde_116}) and $\tilde{\tilde{\xi}}$ (\ref{forall:xitildetilde:117}). We use these maps and flatten them out, row by row, map by map, into the flattend layer $\pi$ (\ref{const_flatten_94}). \\
Explanation of (\ref{const_fullyconnect_relu_95}): we then construct the fully-connected layer $\phi$. Every component of $\phi$ (\ref{forall:iphi_118}) is a linear combination of each flattend layer component together with a connecting weight. Every component $\mathfrak{k}$ of the flattend layer is connected with every component $i_\phi$ of the fully-connected layer with corresponding weight parameters $w_\pi(\mathfrak{k}, i_\phi) \in \mathbb{R}$. Note that the size of the maps $\tilde{\tilde{h}}^{C} \times \tilde{\tilde{w}}^{C}$ is unchanged from the last block. For ReLU activation, we introduce variable $\tilde{s}$ (\ref{const_fullyconnect_relu_96}) and binary activation variable $\tilde{\zeta}$ (\ref{const_fullyconnect_relu_97}). For (\ref{const_fullyconnect_relu_98})-(\ref{const_fullyconnect_relu_99}) we proceed in similar terms to (\ref{const_relu_88})-(\ref{const_relu_89}).\\
Explanation of (\ref{const_output_100}): finally we introduce the output class layer $\psi$ with $n_\psi$ classes (\ref{forall:ipsi_119}), whose weights are fully connected to the previous fully-connected layer. As pointed out in the assumptions, we cannot use sigmoid or softmax as activation functions, but we know how to construct ReLU as linear constraint for this MILP: (\ref{const_output_101})-(\ref{const_output_104}) is an identical approach to the ReLU activation above.\\
Explanation of (\ref{const_bounds_105})-(\ref{const_bounds_112}): similar to (\ref{MILP_model3}), we introduce upper- and lower bounds for each unit in order for modern MILP solvers to work more efficiently:
\begin{align}
&lb(A_\beta^c)(\cdot, \cdot) \ = \ 0 \quad \forall \ c>1, \ \forall \ \beta \\
&lb(B_\delta^c)(\cdot, \cdot) \ = \ 0 \quad \forall \ \delta, \forall \ c \\
&lb(\hat{B}_\delta^c)(\cdot, \cdot) \ = \ 0 \quad \forall \ \delta, \forall \ c \\ 
&lb(\pi)(\cdot) \ = \ 0 \\
&lb(\phi)(\cdot) \ = \ 0 \\
&lb(s_\delta^c)(\cdot, \cdot) \ = \ 0 \quad \forall \ \delta, \ \forall \ c \\
&lb(\tilde{s})(\cdot) \ = \ 0 \\
&lb(\tilde{\tilde{s}})(\cdot) \ = \ 0 \\
\notag \\
&ub(A_\beta^c)(\cdot, \cdot) \ \in \ \mathbb{R}_+ \cup \{+\infty\} \quad \forall \ c>1, \ \forall \ \beta \\
&ub(B_\delta^c)(\cdot, \cdot) \ \in \ \mathbb{R}_+ \cup \{+\infty\} \quad \forall \ \delta, \forall \ c \\
&ub(\hat{B}_\delta^c)(\cdot, \cdot) \ \in \ \mathbb{R}_+ \cup \{+\infty\} \quad \forall \ \delta, \forall \ c \\ 
&ub(\pi)(\cdot) \ \in \ \mathbb{R}_+ \cup \{+\infty\} \\
&ub(\phi)(\cdot) \ \in \ \mathbb{R}_+ \cup \{+\infty\} \\
&ub(s_\delta^c)(\cdot, \cdot) \ \in \ \mathbb{R}_+ \cup \{+\infty\} \quad \forall \ \delta, \ \forall \ c \\
&ub(\tilde{s})(\cdot) \ \in \ \mathbb{R}_+ \cup \{+\infty\} \\
&ub(\tilde{\tilde{s}})(\cdot) \ \in \ \mathbb{R}_+ \cup \{+\infty\}
\end{align}
One way of calculating tight upper bounds is to step through all units: we fix a unit and delete all constraints and variables associated with any other unit in either the same layer or in any higher layer, and then we solve the model (\ref{obj_fct_conv_77})-(\ref{forall:ipsi_119}) in one round to maximize each unit. This gives as a far more accurate tight upper bounds for each unit's output and accelerates MILP solvers.

\subsection{Creating adversarial examples}
The described MILP model in (\ref{CNN_MILP_formulation}) however is not suited for training. We have kernel values $k_\gamma^c(i,j)$ and weight parameters in the fully-connected layer $w_\pi(\mathfrak{k}, i_\phi)$, $w_\phi(\mathfrak{f}, i_\psi)$ to be optimized, but these are \textit{fixed} initializations in (\ref{CNN_MILP_formulation}). We do not have any training elements involved. Instead, the MILP is designed to implicity compute the best possible input example $A_\beta^1(\lambda, \xi) \quad \forall \ \lambda=1, \ldots, h^1, \ \forall \ \xi =1, \ldots, h^1$, that can best be classified by the network.\\
Inversely, we can modify the MILP to compute input examples that are worst possibly classified by the network. This will result in slightly different inputs, called adversarial examples, that the CNN will missclassify upon. This procedure is analogeous to (\ref{DNN:adversarial}) where we found adversarials with DNNs. Again, we will take the MNIST dataset to base the MILP upon. If an image of a digit $A_\beta^1$ is classified correctly as $d$, the goal of our CNN MILP is to find a similar image $\tilde{A}_\beta^1$ which is classified as $\tilde{d}$, with $\tilde{d}\neq d$. As already done by \cite{fischetti17}, one way is to set $\tilde{d} = (d+5) \mod 10$, so the adversarial image of a $3$ should have label $2$.\\
Say we want the activation of the required wrong digit in the output class layer to be at least $20\%$ larger than any other activation, we get
\begin{align} 
\tilde{\psi}(\tilde{d}) \ \geq \ 1.2 \cdot \tilde{\psi}(i_{\tilde{\psi}}) \quad \forall \ i_{\tilde{\psi}}=\{1,\ldots ,10 \} \backslash \tilde{d}, \label{20activation:cnn_144}
\end{align}
where $\tilde{\psi}(\cdot)$ corresponds to the output class layer units of the adversarial.
One can also think of modifying the cost function accordingly
\begin{align}
\min \quad \sum_{i_{\tilde{\psi}}=1}^{10} \tilde{\tilde{c}}(i_{\tilde{\psi}}) \cdot \tilde{\psi}(i_{\tilde{\psi}}) \label{cost_fkt_cnn_adversarial}
\end{align}
with $\tilde{\tilde{c}}(\tilde{d})$ as negative cost: we can encourage the activation of the required wrong digit $\tilde{d}$. Conceivably, we can further penalize high activations of the other units $\tilde{\psi}(i_{\tilde{\psi}}) \ \, \forall \ i_{\tilde{\psi}}=\{1, \ldots, 10 \} \backslash \tilde{d}$ with positive costs.\\
For the adversarial $\tilde{A}_\beta^1$ to be as similar as possible to $A_\beta^1$, we change every image pixel, such that the difference between them is close to $0$:
\begin{align} 
&\min \ \sum_{\beta=1}^{\alpha(A^1)} \sum_{\lambda=1}^{h^1} \sum_{\xi=1}^{w^1} \epsilon_\beta(\lambda, \xi) \label{cost_fkt_cnn_adversarial_146} \\
& -\epsilon_\beta(\lambda, \xi) \ \leq \ A_\beta^1(\lambda, \xi) - \tilde{A}_\beta^1(\lambda, \xi) \ \leq \ \epsilon_\beta(\lambda, \xi) \label{cnn:difference} \\
&\epsilon_\beta(\lambda, \xi) \ \geq \ 0 \\
&\beta=1, \ldots, \alpha(A^1) \label{cnn:adversarial1}\\
&\lambda = 1, \ldots, h^1 \label{cnn:adversarial2} \\
&\xi = 1, \ldots, w^1. \label{cnn:adversarial3}
\end{align}
An additional constraint $\epsilon_\beta(\lambda, \xi) \leq 0.2$ for all (\ref{cnn:adversarial1})-(\ref{cnn:adversarial3}) guarantees that no pixel is changed by more than 0.2. This means that instead of few significantly changed pixels, instead more pixels are changed in total less significantly.\\
By adding (\ref{cost_fkt_cnn_adversarial}) and (\ref{cost_fkt_cnn_adversarial_146}) to the objective function of the MILP, as well as constraints (\ref{cnn:difference})-(\ref{cnn:adversarial3}) and (\ref{20activation:cnn_144}) to the MILP, we can construct such an adversarial example $\tilde{A}_\beta^1$.

\subsection{Evaluation}

Historically CNNs are the most popular form of artificial neural network to perform image application tasks e.g. LeNet-$5$ \cite{yann} as one of the first CNNs by Yann Lecun \cite{yann89}, deep CNN AlexNet \cite{cifar100} used for CIFAR dataset classification and first introduced ReLU activation \cite{cs231:optimizers}, or deep CNN ResNet \cite{resnet50} using network layers to fit a residual mapping \cite{cs231:optimizers}. For good reason, because they all achieved state-of-the-art accuracy at moderate to affordable computational cost \cite{cs231:optimizers}. They maintain translational invariance and shared weight parameters is more efficient than vanilla DNNs.\\
Feeding the MILP model (\ref{CNN_MILP_formulation}) into a modern MILP solver (CPLEX) to create adversarial examples remains to be done. These may then be compared to the results of the DNN MILP presented by \cite{fischetti17}, in order to arrive at an evaluation which MILP model is more suitable.

\newpage
\section{Capsule Networks}
\label{section:capsnet}

This chapter is widely based on \cite{hinton17} by Geoffrey Hinton.

\subsection{Equivariance}
With the help of convolutional kernels, CNNs manage to detect objects in images that are translational invariant (\ref{dog1 image}).The convolutional filters move iteratively through all image areas, which makes it possible to translate the same structure detected in one region across the entire image and keep that information stored within its kernel weights. Training a CNN with a center-positioned object, enables the CNN to classify that same object to the left, right, up or down (\ref{dog1 image}). This is why CNNs are robustly resistant to translational invariance \cite{Goodfellow-et-al-2016}. However, this does not apply for any \textit{transformation} applied to the object in the image. If the object is rotated, flipped, scaled, deformed, reassembled etc. the CNN fails to classify the object correctly \cite{hinton17}. Further training with these transformed objects would be needed and further kernels would be needed to cover all the different transformations. In short: CNN lack \textit{equivariance} \cite{hinton17}. Equivariance is a concept to describe how objects still stay the same, eventhough they are transformed, rotated, in a different light etc. and thus need to be classified as the objects they are, regardless of any transformation \cite{hinton17}. This concept matches how human vision works: when an object is viewed, the human eye sets fixation points to process the area of the object at an high resolution while ignoring irrelevant details \cite{hinton17}. Furthermore, the brain is capable to understand the \textit{instantiation} of a transformed object, in order to identify it.\\
This is the reason and motivation for \cite{hinton17} to introduce a new kind of architecture to classify objects equivariantly - in the following, we follow up on this innovative architecture for computer vision, namely \textit{Capsule Networks} (CapsNets) \cite{hinton17}. CapsNets expand on human vision such that objects are detected as such, regardless of any transformation in the image. Also, in crowded scenes with overlapping elements each element can segmentedly be classified \cite{hinton17}.\\
CapsNets are based on CNNs, replacing max pooling by an \textit{routing-by-agreement} algorithm \cite{hinton17}. Max pooling extracts the major element in a conv map and erases the rest. This causes a lot of information to be lost, moreover it does not serve to achieve equivariance \cite{hinton17}. Maintaining translational invariance however in addition to equivariance is certainly desired for CapsNets.

\begin{figure}[!htb]
\vskip 0.2in
\begin{center}
\centerline{\includegraphics[width=\columnwidth]{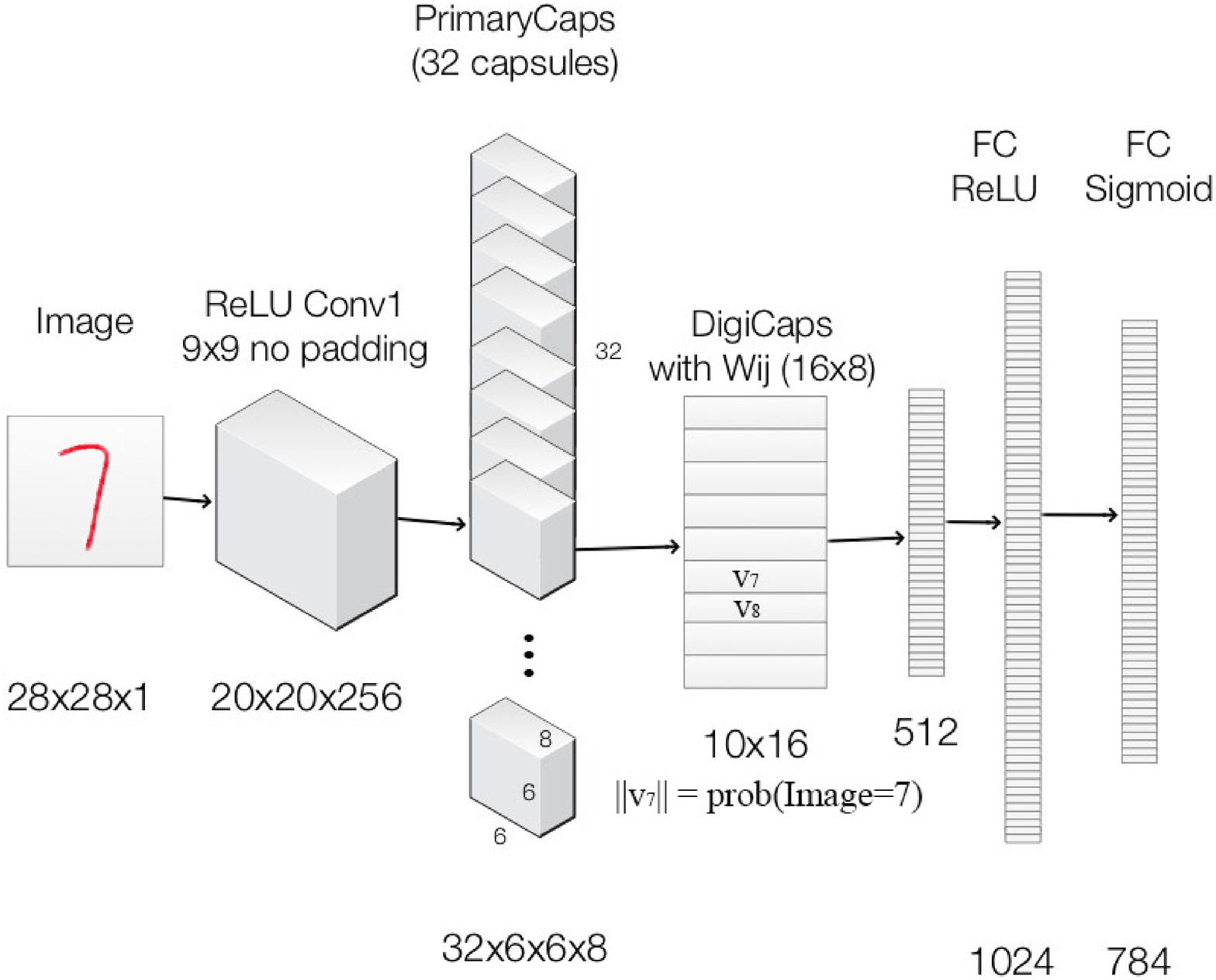}}
\caption{CapsNet Architecture \cite{arch1}\label{arch1}}
\end{center}
\vskip -0.2in
\end{figure}

In the following, we discuss the CapsNet architecture presented in \cite{he15} and our goal is to arrive to an evaluation, comparing CapsNet to state-of-the-art CNNs in image recognition applications.\\
The dataset on which CapsNet in \cite{hinton17} is based on, is the MNIST dataset (\cite{wiki_mnist}).

\subsection{CapsNet Architecture}
\subsubsection{Input layer}

Let the input be a grey-scaled $28 \times 28$ MNIST image i.e. a tensor of integer values ranging between $[0,255]$.

\subsubsection{Convolutional layer}
The first layer is a standard convolutional layer with included ReLU operation. We maintain convolutional layers for CapsNets, because we want to "replicate learned knowledge across space" \citep{hinton17}, thus preserving translational invariance. This first layer of convolutional operations is looking for low level edges and simple curves in the input image. $256$ different $9 \times 9$ convolutional kernels are applied, with a stride of $1$ and no padding. This gives an output of $256 \times [20 \times 20]$ maps.\\
Let $m_\beta$ denote these $256$ maps for $\beta=1, \ldots, 256$, and $m_\beta(k,l)$ be the value in row $k$ and column $l$ of the $\beta^{\text{th}}$ map, $\forall \ \beta=1, \ldots, 256$, $\forall \ k=1, \ldots 20$, $\forall \ l=1, \ldots 20$. ReLU (\ref{relu}) is applied as activation function on all $m_\beta(k,l) \quad \forall \ \beta=1, \ldots, 256$, $\forall \ k=1, \ldots 20$, $\forall \ l=1, \ldots 20$.  So we have $256$ convolutional maps of size $9 \times 9$ consisting of non-negative integers. An additional bias term leads to $(9*9+1)*256$ trainable parameters in this layer \cite{medium:capsnet}.

\subsubsection{PrimaryCaps layer}
So called PrimaryCapsules \cite{hinton17} is a convolutional layer with $32$ "primary capsules" of convolutional $8$-D capsules. Concretely, we have 32 multidimensional "kernels" (or primary capsules), each consisting of $8$ convolutional $9\times9$ kernels and a stride of $2$, that sees the whole input of the $m_1, \ldots, m_{256}$ maps. The result is $8$-D $32 \times [6 \times 6]$ maps. Basically the $256$ input maps are restacked into $32$ "decks" with $8$ maps each deck. This gives us $32$ times $[6 \times 6]$ stacks, each consisting of $8$-D vector components.\\
Let $\tilde{m}_1, \ldots , \tilde{m}_{32}$ be the stacked decks, and $\tilde{m}_{\tilde{\beta}}(\tilde{k},\tilde{l})$ be the vector in row $\tilde{k}$ and column $\tilde{l}$ in the $\tilde{\beta}^{\text{th}}$ deck, $\forall \ \tilde{\beta}=1, \ldots, 32$, $\forall \ \tilde{k}=1, \ldots 6$, $\forall \ \tilde{l}=1, \ldots 6$. Then the $8$-D vector $v=(v_1, \ldots , v_8)$ in $\tilde{m}_{\tilde{\beta}}(\tilde{k},\tilde{l})$ is defined by
\begin{align*}
&v_1=m^\circ_{1+8(\tilde{\beta}-1)}(k,l),\\
&v_2=m^\circ_{2+8(\tilde{\beta}-1)}(k,l),\\
&v_3=m^\circ_{3+8(\tilde{\beta}-1)}(k,l),\\
&v_4=m^\circ_{4+8(\tilde{\beta}-1)}(k,l),\\
&v_5=m^\circ_{5+8(\tilde{\beta}-1)}(k,l),\\
&v_6=m^\circ_{6+8(\tilde{\beta}-1)}(k,l),\\
&v_7=m^\circ_{7+8(\tilde{\beta}-1)}(k,l),\\
&v_8=m^\circ_{8+8(\tilde{\beta}-1)}(k,l)\\
&\forall \ \tilde{\beta}=1, \ldots, 32, \quad \forall \ k=1, \ldots, 20, \quad \forall \ l=1, \ldots, 20,
\end{align*} 
where $m^\circ$ represents the convoluted map of $m$.
Each deck $\tilde{m}_{\tilde{\beta}}$ is called \textit{capsule layer} and each component $\tilde{m}_{\tilde{\beta}}(\tilde{k},\tilde{l})$ is called \textit{capsule}. Clearly each capsule layer has 36 capsules. One can think of an capsule as a group of neurons that collectively produce an \textit{activity vector} with one element for each neuron. On the one hand, vanilla DNNs output a scalar value for each neuron, which illustrates how high the activation of that neuron is, in other words whether a certain entity of an object is present or not \cite{wiki_convnets}. With CapsNets, on the other hand, a capsule outputs an activity vector \cite{hinton17}.\\
Since each of the $32$ primary capsules applies eight $9 \times 9 \times 256$ convolutional kernels to the layer's input, this leads to $32*8*(9*9+1)*256$ trainable parameters in this layer (including a bias term) \cite{medium:capsnet} that are trained by backpropagation.

\paragraph{Activity vector}

Computer graphic programs take instantiation parameters of an object as input and then output the rendered image. Inverse rendering implies taking an image and finding out the instantiation parameters of existing objects.\\
It is the major task of CapsNets to learn instantiation parameters of objects in an image, in order to achieve equivariance \cite{Image_to_Capsules1}. An objects instantiation parameters include exact position, size, deformation, rotation degree, velocity, lighting, albedo, hue, texture etc \cite{Image_to_Capsules1}.
The instantiation parameters are represented by the orientation of the activity vector of a capsule \cite{hinton17}. The length of the activity vector marks the probability that a certain entity exists, analogeous to the activation value of single-scalared DNN neurons \cite{hinton17}.

\subsubsection{DigitCaps layer}

In the following procedure the activity vectors are "squashed" applying a \textit{squashing function} \cite{hinton17}:
\begin{align} \label{squashing_function}
v_j \ = \ \frac{\left\Vert{s_j} \right\Vert_2^2}{1+\left\Vert{s_j} \right\Vert_2^2} \, \frac{s_j}{\left\Vert{s_j} \right\Vert_2}.
\end{align}
The squashing function scales a vector $s_j$ to have length between $0$ and $1$, while maintaining the orientation. It ensures that long vectors get shrunk to a length slightly below 1 and short vectors get shrunk to almost zero length \cite{hinton17}.\\
The squashing function (\ref{squashing_function}) introduces non-linearity to the CapsNet and acts as an activation function: since the length of the activity vector represents the probability that an object exists, it is desired to have vector lengths not exceeding $1$ and not be inferior to $0$.\\
We apply the squashing function to all $8$-D $32 \times [6 \times 6]$ capsules; we obtain normalized activity vectors, the orientation stays identical.\\
\\
Then the $8$-D $ 32 \times [6 \times 6]$ capsules are reshaped into $8$-D $[1 \times 1] \times 1152 $ capsules called $u_i \in \mathbb{R}^{8 \times 1} \, \ \forall \ i=0, \ldots , 1151$, with
\begin{align*}
&u_0 \ = \ \tilde{m}_{1}(1,1)_{sq}\\
&u_2 \ = \ \tilde{m}_{1}(1,2)_{sq}\\
&\vdots \\
&u_6 \ = \ \tilde{m}_{1}(1,6)_{sq}\\
&u_7 \ = \ \tilde{m}_{1}(2,1)_{sq}\\
&\vdots \\
&u_{12} \ = \ \tilde{m}_{1}(2,6)_{sq}\\
&\vdots \\
&u_{36} \ = \ \tilde{m}_{1}(6,6)_{sq}\\
&u_{37} \ = \ \tilde{m}_{2}(1,1)_{sq}\\
\\
&\vdots \\
\\
&u_{1151} \ = \ \tilde{m}_{32}(6,6)_{sq}\\
\end{align*}
where $\tilde{m}_{\tilde{\beta}}(\tilde{k},\tilde{l})_{sq}$ notates the squashed capsule of $\tilde{m}_{\tilde{\beta}}(\tilde{k},\tilde{l}) \quad \forall \ \tilde{\beta}=1, \ldots, 32, \ \, \forall \ \tilde{k}=1, \ldots, 6, \ \, \forall \ \tilde{l}=1, \ldots, 6$. \\
\\
Now each capsule $u_i$ is multiplied with an individual weight matrix $W_{ij} \in \mathbb{R}^{16 \times 8}$
\begin{align} \label{u_hat}
\hat{u}_{j|i} \ = \ W_{ij} \cdot u_i,
\end{align}
with $u_i$ as input capsule $\forall \ i=0, \ldots, 1151$ and $j = 0, \ldots , 9$ the number of digit classes. 
The weight matrix is an affine transformation matrix and stores learnable information about the exact part-whole relationship of each entity of the entire object. It is initialized at the beginning and the values are learned with backprogagation iteratively. Each capsules activity vector stores instantiation parameters of an objects entity and the weight matrix stores data for the exact spatial relationship of the entity regarding the entire object. E.g. if the input image is a face and capsule $u_i$ stores information about the eye, then the weight matrix gives information on how exactly the eye is to be spatially positioned in the face, such that essentially a proper face is detected. In CNNs, such spatial information is not stored, which is why Pablo Picassos famous "Portrait of woman in d`hermine pass" figure (\ref{picasso_face}) will falsely be classified by CNNs as a face. This makes CNNs vulnerable to adversarial attacks.\\
The matrices $W_{ij}$ represent a part-whole relationship. This tells us in what manner a lower level entity, a capsule $u_i$, fits in to with a higher level entity/object. E.g. $u_i$ may store information about a curve, and $W_{i0}$ represents in what posture curves fit into the digit "0", then $W_{i0} \, u_i = \hat{u}_{0|i}$ has significant length; whereas there are no curves in the digit "1", consequently $W_{i1} \, u_i = \hat{u}_{1|i}$ is a short activity vector for that same curve capsule $u_i$.

\begin{figure}[!htb]
\vskip 0.2in
\begin{center}
\centerline{\includegraphics[width=\columnwidth]{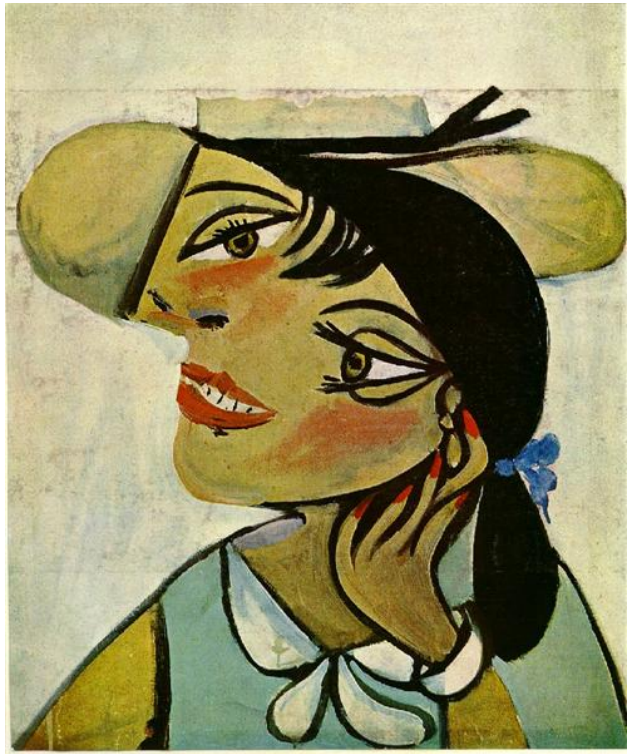}}
\caption{A CNN recognizes two eyes, a nose, a mouth etc. and falsely predicts the class of a face, because it fails to relate the entities in a spatial context. \\ "Portrait of woman in d`hermine pass"\cite{picasso_face}\label{picasso_face}}
\end{center}
\vskip -0.2in
\end{figure}

As in (\ref{u_hat}), for all $i=0, \ldots, 1151$ every capsule $u_i$ is transformed into $\hat{u}_{j|i} \in \mathbb{R}^{16 \times 1}$ via matrix-vector multiplication\footnote{This works, since $W_{ij} \in \mathbb{R}^{16 \times 8}$ and $u_i \in \mathbb{R}^{8 \times 1}$.} with every digit class $j = 0, \ldots ,9$.
The result is  $16$-D $10 \times [1152 \times 1]$  capsules , see figure (\ref{capsnet_visualize}), noted as
\begin{align*}
\hat{u}_{0|0}, &\ldots , \hat{u}_{0|1151} \quad \text{for digit} \ \, 0 \\
\hat{u}_{1|0}, &\ldots, \hat{u}_{1|1151} \quad \text{for digit} \ \, 1 \\
& \vdots \\
\hat{u}_{9|0}, &\ldots, \hat{u}_{9|1151} \quad \text{for digit} \ \, 9.
\end{align*}
So we have $1152$ $16$-D capsules for each of the ten digits.\\
Each of the $1152$ $u_i$ capsules is multiplied with an own $16 \times 8$ matrix $W_{ij}$, so we have $1152*16*8$ trainable weight parameters here, that need to trained using backpropagation.\\
The concept that higher level capsule layers produce higher dimensional capsules matches the idea of higher level convolutional filters producing more complex forms and shapes. High dimensional capsules can store considerably more instantiation parameters for a more complex entity of the object. Finally the last layer of capsules, DigiCaps, represent ten capsules for each class to be detected. They are created by using routing-by-agreement algorithm.

\begin{figure*}[!htb]
\vskip 0.2in
\begin{center}
\centerline{\includegraphics[width=\textwidth]{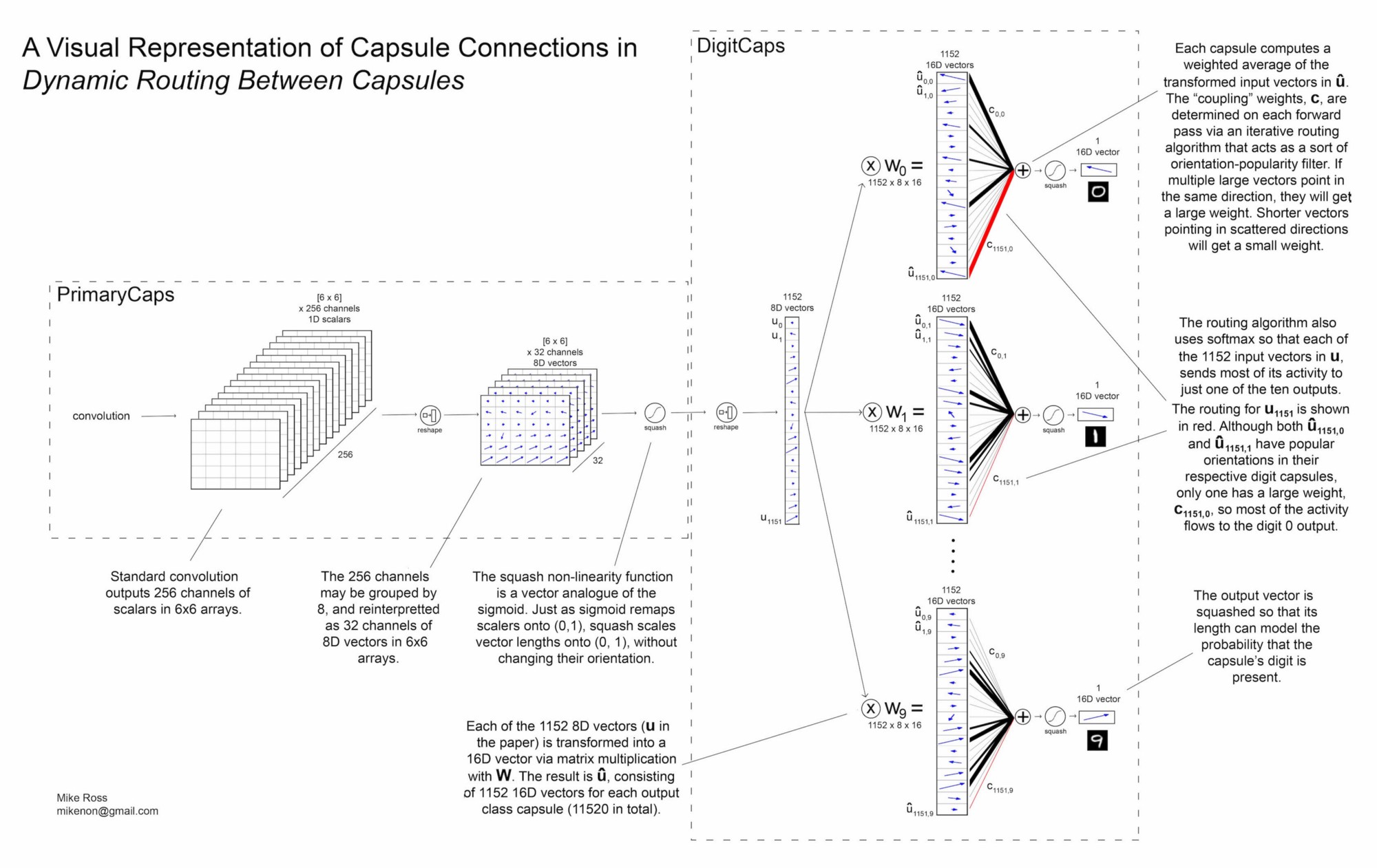}}
\caption{A visualization of PrimaryCaps and DigitCaps \cite{medium:visualpic} \\ Caution, this figure has an error: $\hat{u}_{i|j}$ instead of $\hat{u}_{j|i}$ }\label{capsnet_visualize}
\end{center}
\vskip -0.2in
\end{figure*}

\paragraph{Routing-by-agreement algorithm} 
\label{routing_by_agreement}

For each $j=0, \ldots, 9$ the $16$-D capsules $\hat{u}_{j|0}, \ldots , \hat{u}_{j|1151}$ are sumed up in a way that highly significant capsules with good predictions are weighted profoundly, and less significant capsules have less influence on the next capsule layer \cite{hinton17}. Hinton's paper \cite{hinton17} introduces a routing by agreement algorithm, that iteratively sets these weights, $c_{ij}$, and serves as a better way than max pooling to prioritize the dominant features, while at the same time preserving the less relevant data. 

\begin{algorithm}[!htb]
   \caption{Routing algorithm \cite{hinton17}}
   \label{alg:routing}
\begin{algorithmic}
   \STATE {\bfseries Input:} $\hat{u}_{j|i}$, iteration $r$, layer $l$
   \item for all capsule $i$ in layer $l$ and capsule $j$ in layer $l+1$: $\textbf{b}_{ij} \leftarrow 0$.
   \FOR{ $r$ iterations }
   \item for all capsule $i$ in layer $l$: $\textbf{c}_i$ $\leftarrow$ $\texttt{softmax}(\textbf{b}_i)$
   \item for all capsule $j$ in layer $l+1$: $\textbf{s}_j \leftarrow \sum_i c_{ij} \, \hat{\textbf{u}}_{j|i}$
   \item for all capsule $j$ in layer $l+1$: $\textbf{v}_j \leftarrow  \texttt{squash}(\textbf{s}_j)$
   \item for all capsule $i$ in layer $l$ and capsule $j$ in layer $l+1$: $b_{ij} \leftarrow b_{ij} + \hat{\textbf{u}}_{j|i} \cdot \textbf{v}_j$
   \ENDFOR
   \item \textbf{return} $\textbf{v}_j$
   \newline
   \item with: $\texttt{softmax}(\textbf{b}_i) = \frac{\exp(b_{ij})}{\sum_k \exp (b_{ik})}$
\end{algorithmic}
\end{algorithm}

For every capsule $u_i$ in layer $l$ we have the prediction $\hat{u}_{j|i}$ available, which is calculated by (\ref{u_hat}).\\
On the one hand, if the entity represented of capsule $u_i$ in layer $l$ is not related in any way to the higher level entity of capsule $v_j$ in layer $l+1$, then $\hat{u}_{j|i}$ will only have marginal impact, thus it is the goal of (\ref{alg:routing}) to assign only marginal $c_{ij}$.\\
On the other hand, if many capsules $u_i$ in layer $l$ have similar $\hat{u}_{j|i}$ for capsule $v_j$ in layer $l+1$, then the corresponding $c_{ij}$ will be major, because the capsules agree on what the object or entity of an object looks like, thus there is no need to send large weight $c_{ij}$ to any other capsule. This would only cause noise.\\
With their calculation $\hat{u}_{j|i}$ the capsules $u_i$ in layer $l$ try to predict the capsule $v_j$ in layer $l+1$.

\subsubsection{Example}
\label{example}

In the following we follow up an examplel, introduced by \cite{Image_to_Capsules1}, to better understand the dynamic routing between capsule.\\
Suppose a boat image such as in (\ref{Image_to_Capsules1}) is the input to CapsNet. As we see in (\ref{CapsuleLayer1_to_CapsuleLayer2}), a triangle capsule $u_1$ (blue) and a rectangle capsule $u_2$ (black) become active. The corresponding activity vectors are present, the orientation marks the instantiation of that entity and the length marks the probability. Note that other image areas will result in short vectors for these two specific capsules, we omit these in the framework for the sake of simplicity. 

\begin{figure}[!htb]
\vskip 0.2in
\begin{center}
\centerline{\includegraphics[width=\columnwidth]{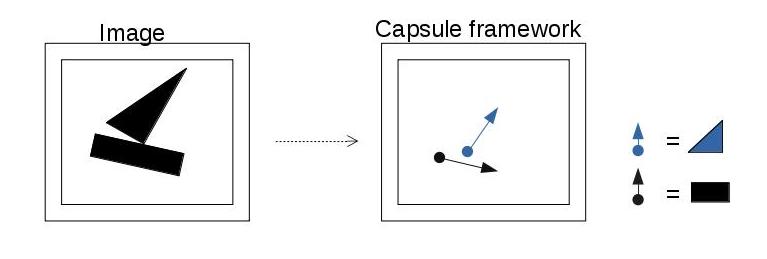}}
\caption{\cite{Image_to_Capsules1}\label{Image_to_Capsules1}}
\end{center}
\vskip -0.2in
\end{figure}

\begin{figure}[!htb]
\vskip 0.2in
\begin{center}
\centerline{\includegraphics[width=\columnwidth]{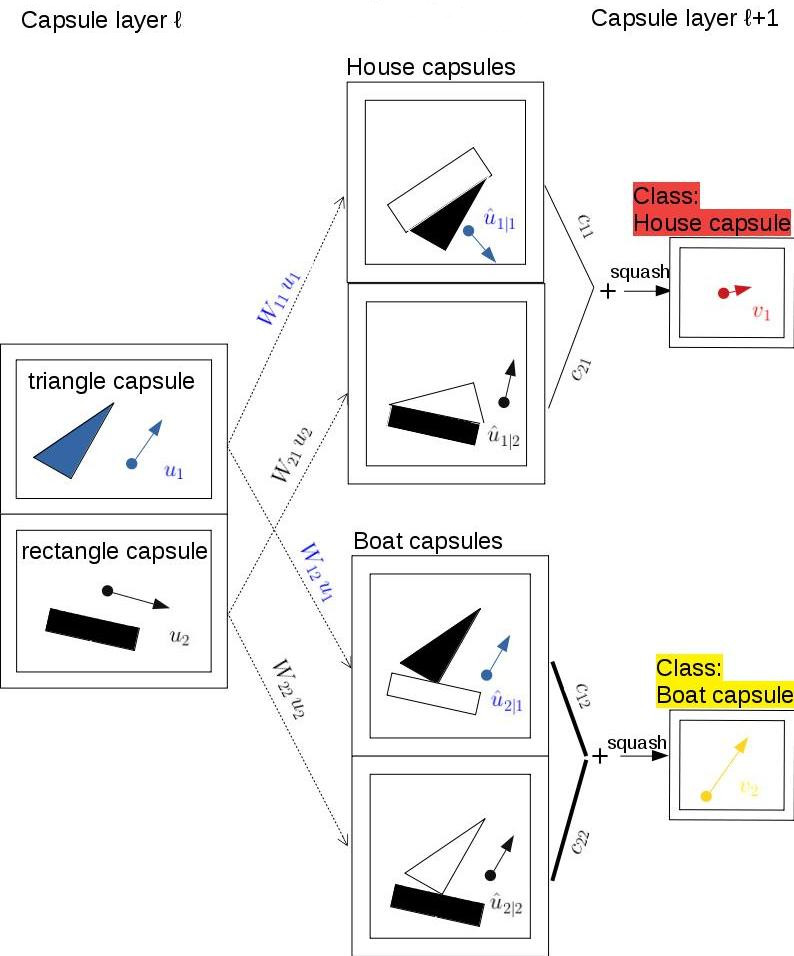}}
\caption{Visualization of routing-by-agreement for Example (\ref{example})   \cite{Image_to_Capsules1}\label{CapsuleLayer1_to_CapsuleLayer2}}
\end{center}
\vskip -0.2in
\end{figure}

Suppose the capsule class layer $l+1$ consists of only two capsule classes: one capsule $v_1$ identifying houses, the other $v_2$ identifying boats.\\
For $u_1$:\\
According to the orientation of $u_1$, the house capsule $v_1$ in layer $l+1$, will predict an upside-down house, this is stored in $\hat{u}_{1|1}$. The information explaining the way the roof fits to the house is stored in the weight matrix $W_{11}$.\\
On the other hand, the boat capsule $v_2$ predicts a slightly tilted boat for $u_1$, this is stored in $\hat{u}_{2|1}$. The information explaining the way the sail fits to the boat is stored in $W_{12}$.\\
For $u_2$:\\
According to the orientation of $u_2$, the house capsule $v_1$ will recognize a slightly tilted house, this is stored in $\hat{u}_{1|2}$. Whereas the boat capsule $v_2$ recognizes a slightly tilted boat from $u_2$, this is stored in $\hat{u}_{2|2}$.\\
It is evident that the capsules in layer $l$ strongly agree on what a boat should look like ($\hat{u}_{2|1}$ and $\hat{u}_{2|2}$ match) and strongly disagree on what a house should look like ($\hat{u}_{1|1}$ and $\hat{u}_{1|2}$ differ). Therefore it is quite likely that the unfamiliar object is essentially a boat. Both capsules in layer $l$ should therefore send most of their output to the boat capsule $v_2$ in layer $l+1$ and only little to the house capsule $v_1$.\\
Concretely, by applying the (\ref{alg:routing}) routing-by-agreement algorithm, we can calculate appropriate weights $c_{ij}$ in order to distribute high weights to well predicting capsules and low weights to poor predicting capsules.\\
We can think of $u_1$ and $u_2$ to be the activity vectors shown in figure (\ref{CapsuleLayer1_to_CapsuleLayer2}).
Calculate each $\hat{u}_{j|i}$:
\begin{align*}
&\hat{u}_{1|1} \ = \ W_{11} \, u_1 \\
&\hat{u}_{1|2} \ = \ W_{21} \, u_2 \\
&\hat{u}_{2|1} \ = \ W_{12} \, u_1 \\
&\hat{u}_{2|2} \ = \ W_{22} \, u_2,
\end{align*}
which is the input to algorithm (\ref{alg:routing}).
We can think of the $\hat{u}_{1|1}, \ \hat{u}_{1|2}, \ \hat{u}_{2|1}, \ \hat{u}_{2|2}$ to be the activity vectors shown in figure (\ref{CapsuleLayer1_to_CapsuleLayer2}).
Following the first step, we receive
\begin{align*}
& b_{11} \ = \ 0 \\
& b_{12} \ = \ 0 \\
& b_{21} \ = \ 0 \\
& b_{22} \ = \ 0.
\end{align*}
\cite{hinton17} has shown that the number of iterations can be $r \in \{ 2, 3 \}$ to be working adequately.
In the first iteration $r=1$, applying softmax (\ref{softmax})\cite{wiki_softmax} gives us
\begin{align*}
& c_{11} \ = \ \frac{\exp(b_{11})}{\exp(b_{11}) + \exp(b_{12})} \ = \ \frac{1}{2} \\
& c_{12} \ = \frac{\exp(b_{12})}{\exp(b_{11}) + \exp(b_{12})} \ \, = \ \frac{1}{2} \\
& c_{21} \ = \ \frac{\exp(b_{21})}{\exp(b_{21}) + \exp(b_{22})} \ = \ \frac{1}{2} \\
& c_{22} \ = \ \frac{\exp(b_{22})}{\exp(b_{21}) + \exp(b_{22})} \ = \ \frac{1}{2}.
\end{align*}
Softmax is similiar to the sigmoid function: delegate value between $0$ and $1$ and all values sum up to $1$.  
It must be $\sum_i c_{ij} = 1$ (capsule $i$ in layer $l$) for all capsules $j$ in layer $l+1$ i.e. $\sum_{i=1}^2 c_{i1} =1$ and $\sum_{i=1}^2 c_{i2} =1$.\\
The next step is calculating $s_j$:
\begin{align} 
& s_1 \ = \ \sum_i c_{i1} \, \hat{u}_{1|i} \ = \ \frac{1}{2} \, \hat{u}_{1|1} + \frac{1}{2} \, \hat{u}_{1|2} \label{calculating_s1}\\
& s_2 \ = \ \sum_i c_{i2} \, \hat{u}_{2|i} \ = \ \frac{1}{2} \, \hat{u}_{2|1} + \frac{1}{2} \, \hat{u}_{2|2}. \label{calculating_s2}
\end{align}
Based on the vectors in figure (\ref{CapsuleLayer1_to_CapsuleLayer2}), we can literally image how $s_1$ and $s_2$ look like:\\
$s_1$ must be a short vector pointing at $2$ o'clock, while $s_2$ must be a longer vector pointing at $1$ o'clock.\\
By squashing $s_1$ and $s_2$, we receive $v_1$ and $v_2$. While maintaining the orientation, $v_1$ is now even shorter and $v_2$ is longer.\\
The $b_{ij}$ can now be updated according to how much each $\hat{u}_{j|i}$ agrees with $v_j$. We use the dot product to measure this agreement (marked by "$\circ$"), because the dot product of two vectors depends on their length and in what angle they relate to each other \cite{wiki:dotproduct}.
\begin{align*}
& b_{11} \ = \ b_{11} + \hat{u}_{1|1} \circ v_1 \\
& b_{12} \ = \ b_{12} + \hat{u}_{2|1} \circ v_2 \\
& b_{21} \ = \ b_{21} + \hat{u}_{1|2} \circ v_1 \\
& b_{22} \ = \ b_{22} + \hat{u}_{2|2} \circ v_2.
\end{align*}
Now $b_{12}$ and $b_{22}$ have much greater value.
In iteration $r=2$ and $r=3$, the weights $c_{12}$ and $c_{22}$ will be rated even higher, since they will have greater impact on (\ref{calculating_s1}),(\ref{calculating_s2}).\\
Finally, the next-layer capsules, $v_j$ of layer $l+1$ ,are returned: a short red $v_1$ capsule and long yellow $v_2$ capsule (\ref{CapsuleLayer1_to_CapsuleLayer2}). These capsules represent the best over-all predictions made by all the primary capsules in layer $l$. \\
For each class capsule $v_j$, there are $1152$ $16 \times 8$ weight matrices. In addition, the routing-by-agreement algorithm needs to $1152$ variables for $c_i$ and another $1152$ variables for $b_i$. For $10$ class capsules, this sums up to $\left( 1152 * 16 * 8 + 1152 * 2 \right) * 10$ trainable parameters in the Digit Caps layer - the weight matrices are learned by backpropagation, the routing parameters are learned by routing-by-agreement.
\\
\\
Following up on (\ref{routing_by_agreement}), the $11520 \times 16$D capsules are sumed up in the manner of $\hat{u}_{j|1}, \ldots , \hat{u}_{j|1152} \ \, \forall \ j=0, \ldots, 9$ such that each of the weights $c_{ij}$ is properly set according to the agreement $\hat{u}_{j|i} \cdot \text{squash}(\sum_i c_{ij} \hat{u}_{j|i})$. Routing by agreement delivers not only suitable weights $c_{ij}$, but also returns $v_j \ \forall \ j=0,\ldots, 9$ - the result are $10 \times 16$D squashed class capsules $v_j$, $ j=0, \ldots, 9 $ for each class digit.

\subsubsection{Margin Loss}

To allow the prediction of multiple classes, we use a separate margin loss, $L_k$ for each digit capsule $k$:
\begin{align}
L_k \ &= \ T_k \ \max  (0, m^+ - \lVert v_k \rVert ) ^2 \, \notag \\
&+ \, \lambda \, (1- T_k) \, \max (0, \lVert v_k \rVert - m^- )^2.
\end{align}
with $T_k = 1$ if and only if digit of class $k$ is present and $m^+ = 0.9$, $m^- = 0.1$ and $ \lambda = 0.5$.\\
If digit $k$ is on the input image and corresponding digit capsule $v_k$ ends up having longer length than $m^+$, then we have no margin loss. If digit $k$ is not on the input image and corresponding digit capsule $v_k$ ends up being shorter than $m^-$, then we have no margin loss. With $\lambda$ we can regularize the loss for falsely classified digits.

\subsubsection{Reconstruction as a regularization method}

As \cite{hinton17} proposes, we can add a reconstruction loss to encourage the digit capsules to construct the instantiation parameters of the input image. During training, all but the true activity vector are blocked out and this activity vector is then used to reconstruct the input image. This reconstruction system is called decoder \cite{hinton17} and consist of $3$ fully-connected layers attached to DigiCaps, see figure (\ref{decoder_pic}). The true activity vector is fed into the decoder and the final layer consists of $28*28$ units representing the pixel values of the reconstructed image. We can add the minimization of the squared difference between the input image pixel values and the reconstructed image pixel values to the sum of margin losses, and receive a total loss function. The reconstruction loss is down-scaled by $0.0005$ \cite{hinton17}, so that the margin loss is the dominating factor. We use backpropagation and a gradient descent optimizer (adam) to minimize the sum of the margin losses $\sum_k L_k$ plus the squared error of the decoder \cite{hinton17}.\\

\begin{figure}[!htb]
\vskip 0.2in
\begin{center}
\centerline{\includegraphics[width=\columnwidth]{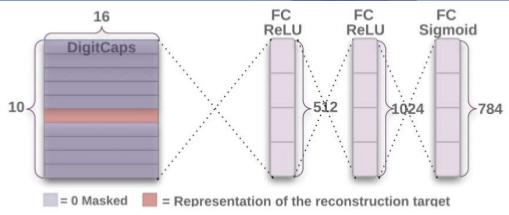}}
\caption{Decoder structure to reconstruct image from a digit capsule \cite{hinton17}}\label{decoder_pic}
\end{center}
\vskip -0.2in
\end{figure}

\begin{figure}[!htb]
\vskip 0.2in
\begin{center}
\centerline{\includegraphics[width=\columnwidth]{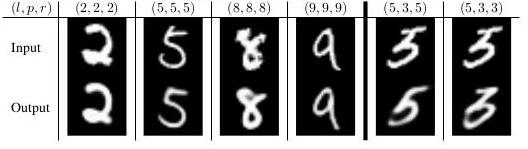}}
\caption{ \cite{hinton17} Reconstruction examples of MNIST test dataset with $3$ routing iterations. $(l,p,r$ stands for the label, the prediction of the CapsNet and the reconstruction respectively. The two far right columns represent two failure examples. The other columns with correct classification show that the reconstructions perpetuate many details and cancel out noise.}
\end{center}
\vskip -0.2in
\end{figure}

\subsection{CapsNet Training}

An image is fed into the Input Layer. Applying a convolutional layer leads to various conv filters that represent certain low level shapes and angles in the image. These filters are stacked to capsule layers and squashed, in which each capsule represents a low level entity of the image. The activity vectors of each capsule store the instantiation parameters of the entity, the length shows the probability of it existing. The capsules associated to the digit in the image have a striking longer length than the others. In what manner each of entity is part of a whole class i.e. in what spatial relationship each entity is connected to the whole object, is defined by the weight matrix $W_{ij}$. Each capsule is then multiplied with a weight matrix $W_{ij}$. The result is $\hat{u}_{j|i}$, again in which the well predicting capsules have long length. Then routing by agreement sets the weights $c_{ij}$ according to how much $\hat{u}_{j|i}$ agrees with the mean $ \text{squash}(\sum_i c_{ij} \hat{u}_{j|i})$ i.e. the well predicting capsules receive a higher weight and contribute greater than others. The algorithm returns a Digit Caps layer, namely $10 \times 16$D squashed capsules $v_0, \ldots , v_9$. These high level capsules store high level features of more complex form, concretely, the entire digit. The result of an input image is that the corresponding capsule has long length and the other capsules have short length. The length characterizes the probability of a digit in the image and yields a prediction.\\
As \cite{hinton17} pointed out, one can analyze individual dimensions of a capsule to learn of what instantiation it is represented by. There are programs, such as \cite{capsule_perturbation}, that perturb the dimensions of a capsule, making visible which instantiation is associated with the stroke thickness, skew, width etc. This might help improve transparency and accountability, since each capsules instantiation in every capsule layer can be exposed \cite{explainability}.\\
A tensorflow implementation of CapsNet for MNIST can be found at
\url{https://github.com/MJimitater/CapsNet/blob/master/CapsNet.ipynb}.

\subsection{Performance and Evaluation}

The number of trainable parameters (including bias units) in CapsNet presented for the MNIST dataset in \cite{hinton17}, figure (\ref{arch1}), are:
\begin{itemize}
\item Convolutional layer: \\
$(9*9+1)*256=20,992$ 
\item PrimaryCaps layer: \\
$(32*8)*((9*9+1)*256)=5,373,952$
\item DigitCaps layer: \\
$(1152*(16*8) + 1152*2))*10 = 1,497,600$ 
\begin{itemize}
\item sum (without decoder): $6,892,544$
\end{itemize}
\item Decoder: $1^{\text{st}}$ fully-connected layer: $(512+1)*1024$
\item Decoder: $2^{\text{nd}}$ fully-connected layer:
$(1024+1)*784$
\begin{itemize}
\item sum (with decoder): $8,221,456$
\end{itemize}
\end{itemize}

Other than a slight pixel shift, no data augmentation is performed in Hinton's paper \cite{hinton17}. The baseline to compare is a three convolutional layer CNN with $256, 256, 128$ kernels of size $5\times5$ and stride of $1$. The last layer is followed by two fully-connected layers of size $328, 192$, and finally the output class layer of $10$ units, softmax activation, dropout, and adam optimizer \cite{hinton17}. The baseline sports $35.4$ million trainable parameters, however the paper \cite{hinton17} is not entirely clear on how this number is calculated. The major property is to design the baseline to have best possible performance on MNIST wile keeping computation cost on a similar level to CapsNet.

\begin{figure}[!htb]
\vskip 0.2in
\begin{center}
\centerline{\includegraphics[width=200pt]{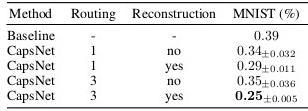}}
\caption{MNIST average error rate and standard deviation achieved by \cite{hinton17}. CapsNet is able to get a lower error rate than the baseline.}\label{MNIST_performance}
\end{center}
\vskip -0.2in
\end{figure}

The CapsNet's architecture is the same as the one disussed in (\ref{arch1}). As figure (\ref{MNIST_performance}) of \cite{hinton17} shows, CapsNet scores lower error rates than the baseline. The presented results also show the importance of the reconstruction decoder. It enforces the object's instantation to be "encapsulated". The number of routing iterations also has an effect: a major positive effect with decoder, a minor negative effect without decoder.\\
Eventhough \cite{hinton17} mentions notable performance of CapsNet on the MultiMNIST dataset and on the CIFAR10 dataset, concretely that the error rates match the ones of when standard CNNs first were applied to these datasets. CapsNet is a new artificial neural network architecture and is subject to further improvements to come.\\
\\
\cite{CapsNet_test_eval} gives insights to more extensive performance testing of CapsNet. The CapsNet as in figure (\ref{arch1}), excluding the decoder subnetworks, was applied to several large datasets, such as Yale Face Database B \cite{yale} with $38$ classes, BelgiumTS traffic signs dataset \cite{belgium} with $62$ classes and CIFAR-$100$ \cite{cifar100} with $100$ classes, among others.\\
As baseline, different CNNs were used, including modified LeNet \cite{yann} and Resnet $50$ \cite{resnet50}.\\
On these large datasets with many more classes than MNIST, the baseline CNNs outperform CapsNet in both accuracy and average training time using GPU computing power \cite{CapsNet_test_eval}.

\begin{figure}[!htb]
\vskip 0.2in
\begin{center}
\centerline{\includegraphics[width=\columnwidth]{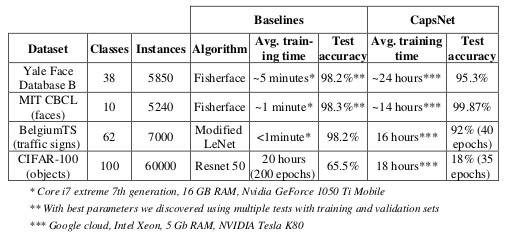}}
\caption{ \cite{CapsNet_test_eval} The baseline CNNs train a lot faster and score higher accuracy than CapsNet almost every time.  }
\end{center}
\vskip -0.2in
\end{figure}

Regarding computational cost, CapsNet requires a lot more than CNNs. This is due to the fact that capsules output higher dimensional activity vectors with instantiation parameters than just scalar products, causing greater GPU memory usage.\\
There are three drawbacks presented by \cite{CapsNet_test_eval}:\\
First of all: for complex images, even higher dimensional capsules are needed to store all instantiation parameters, $32$-D or even $64$-D capsules. This would need further powerful GPUs with more memory to fit the capsule sizes.\\
Second of all: CapsNets achieve good accuracy with small datasets of simple complexity, but fail to hold up with CNNs at large complex datasets. It turns out that on complex datasets CapsNet also need many training images, this equalizes the hope for needing less training images. This hope only holds for simple and less complex images, as researched by \cite{lung_cancer}. CNNs do possess undesirable properties e.g. lack of equivariance, however, on complex datasets, training with transformed images can solve this \cite{wiki_convnets}.\\
Third of all: increasing the size of the input images drastically increases computational cost. Downsizing is only a limited option, since it implies loss of information.\\
\\
As investigated by \cite{brain_tumor}, modified CapsNets perform decently better than baseline CNNs in MRI brain tumor classification, as capsules handle small datasets well, despite the complexity in the images. Equivariance helps CapsNet to classify brain tumors more accurately than CNNs.\\
We can adhere that the success of CapsNets depend on the specific image application. It is yet to be researched in how far the image complexity matters, as well as the number of classes, the number of capsule layers and other hyperparameters.\\ 
As \cite{CapsNet_test_eval} points out, it becomes clear that CapsNets are at an early stage of development and lack scalability, as CNNs were too when they first were applied \cite{wiki_convnets}. Further improvements are needed to boost accuracy and ameliorate computational cost, like done by \cite{lung_cancer}. Several modifications have already been introduced by Hinton in an updated paper \cite{emrouting} e.g. matrix capsules which group capsules to form part-whole relationships, rather than having weight matrices for this. Further testing and modification of hyperparameters is needed for CapsNet to utilize its full potential.\\

\newpage

\bibliography{thesis_template1}

\begin{thebibliography}{40}
\providecommand{\natexlab}[1]{#1}
\providecommand{\url}[1]{\texttt{#1}}
\expandafter\ifx\csname urlstyle\endcsname\relax
  \providecommand{\doi}[1]{doi: #1}\else
  \providecommand{\doi}{doi: \begingroup \urlstyle{rm}\Url}\fi

\bibitem[Afshar et~al.(March, 2018)Afshar, Mohammadi, and
  Plataniotis]{brain_tumor}
Afshar, Parnian, Mohammadi, Arash, and Plataniotis, Konstantinos~N.
\newblock Brain tumor type classification via capsule networks.
\newblock March, 2018.
\newblock URL \url{https://arxiv.org/pdf/1802.10200.pdf}.

\bibitem[Avrutskiy(December, 2017)]{avrutskiy17}
Avrutskiy, V.I.
\newblock Backpropagation generalized for output derivatives.
\newblock December, 2017.
\newblock URL \url{https://arxiv.org/pdf/1712.04185.pdf}.

\bibitem[Bourdakos(February, 2018)]{understanding_kernels}
Bourdakos, Nick.
\newblock Understanding capsule networks - ai's alluring new architecture.
\newblock February, 2018.
\newblock URL
  \url{https://medium.freecodecamp.org/understanding-capsule-networks-ais-\
  alluring-new-architecture-bdb228173ddc}.

\bibitem[Brownlee(September, 2016)]{performance_tweaks}
Brownlee, Jason.
\newblock How to improve deep learning performance.
\newblock September, 2016.
\newblock URL
  \url{https://machinelearningmastery.com/improve-deep-learning-performance/}.

\bibitem[Chevalyre(2017)]{chevalyre}
Chevalyre, Yann.
\newblock Cours data mining / machine learning by yann chevalyre, paris
  dauphine université.
\newblock 2017.

\bibitem[Dauphin et~al.(June, 2014)Dauphin, Pascanu, Gulcehre, Cho, Ganguli,
  and Bengio]{local_minima}
Dauphin, Yann~N., Pascanu, Razvan, Gulcehre, Caglar, Cho, Kyunghyun, Ganguli,
  Surya, and Bengio, Yoshua.
\newblock Identifying and attacking the saddle point problem in
  high-dimensional non-convex optimization.
\newblock June, 2014.
\newblock URL \url{https://arxiv.org/pdf/1406.2572.pdf}.

\bibitem[Fischetti \& Jo(December, 2017)Fischetti and Jo]{fischetti17}
Fischetti, Matteo and Jo, Jason.
\newblock \emph{Deep Neural Networks as 0-1 Mixed Integer Linear Programs: A
  Feasibility Study}.
\newblock December, 2017.
\newblock URL \url{https://arxiv.org/pdf/1712.06174.pdf}.

\bibitem[Goodfellow et~al.(2016)Goodfellow, Bengio, and
  Courville]{Goodfellow-et-al-2016}
Goodfellow, Ian, Bengio, Yoshua, and Courville, Aaron.
\newblock \emph{Deep Learning}.
\newblock MIT Press, 2016.
\newblock \url{http://www.deeplearningbook.org}.

\bibitem[Géron(2017)]{Image_to_Capsules1}
Géron, Aurélien.
\newblock Capsule networks (capsnets) – tutorial.
\newblock 2017.
\newblock URL \url{https://www.youtube.com/watch?v=pPN8d0E3900&t=434s}.

\bibitem[He et~al.(December, 2015)He, Zhang, Ren, and Sun]{resnet50}
He, Kaiming, Zhang, Xiangyu, Ren, Shaoqing, and Sun, Jian.
\newblock Deep residual learning for image recognition.
\newblock December, 2015.
\newblock URL \url{https://arxiv.org/pdf/1512.03385.pdf}.

\bibitem[He et~al.(February, 2015)He, Zhang, Ren, and Sun]{he15}
He, Kaiming, Zhang, Xiangyu, Ren, Shaoqing, and Sun, Jian.
\newblock \emph{Delving Deep into Rectifiers: Surpassing Human-Level
  Performance on ImageNet Classification}.
\newblock February, 2015.
\newblock URL \url{https://arxiv.org/pdf/1502.01852.pdf}.

\bibitem[Hinton et~al.(May, 2018)Hinton, Sabour, and Frosst]{emrouting}
Hinton, Geoffrey, Sabour, Sara, and Frosst, Nicholas.
\newblock Matrix capsules with em routing.
\newblock May, 2018.
\newblock URL \url{https://openreview.net/pdf?id=HJWLfGWRb}.

\bibitem[Hinton et~al.(2017)Hinton, Sabour, and Frosst]{hinton17}
Hinton, Geoffrey~E., Sabour, Sara, and Frosst, Nicholas.
\newblock Dynamic routing between capsules.
\newblock 2017.

\bibitem[Hsu(November, 2017)]{capsule_perturbation}
Hsu, Jeremy.
\newblock A partial implementation of the capsule network from dynamic routing
  between capsules.
\newblock November, 2017.
\newblock URL \url{https://github.com/JeremyCCHsu/CapsNet-tf}.

\bibitem[Hui(2017)]{arch1}
Hui, Jonathan.
\newblock Understanding dynamic routing between capsules (capsule networks).
\newblock 2017.

\bibitem[Janocha \& Czarnecki(February, 2017)Janocha and
  Czarnecki]{lossfunctions}
Janocha, Katarzyna and Czarnecki, Wojciech~Marian.
\newblock On loss functions for deep neural networks in classification.
\newblock February, 2017.
\newblock URL \url{https://arxiv.org/pdf/1702.05659.pdf}.

\bibitem[Kaggle(2014)]{kaggle}
Kaggle.
\newblock Cats vs. dogs.
\newblock 2014.
\newblock URL \url{https://www.kaggle.com/c/dogs-vs-cats}.

\bibitem[Karpathy \& Johnson(Spring, 2018)Karpathy and
  Johnson]{cs231:optimizers}
Karpathy, Andrej and Johnson, Justin.
\newblock Cs231n convolutional neural networks for visual recognition.
\newblock Spring, 2018.
\newblock URL \url{http://cs231n.github.io/convolutional-networks/}.

\bibitem[Krizhevsky(2009)]{cifar100}
Krizhevsky, Alex.
\newblock Learning multiple layers of features from tiny images.
\newblock 2009.
\newblock URL \url{https://www.cs.toronto.edu/~kriz/cifar.html}.

\bibitem[Lecun et~al.(1989)Lecun, Boser, Denker, Henderson, Howard, Hubbard,
  and Jackel]{yann89}
Lecun, Yann, Boser, B., Denker, J.S., Henderson, D., Howard, R.E., Hubbard, W.,
  and Jackel, L.D.
\newblock \emph{Handwritten Digit Recognition with a Back-Propagation Network}.
\newblock 1989.
\newblock URL \url{http://yann.lecun.com/exdb/publis/pdf/lecun-90c.pdf}.

\bibitem[LeCun et~al.(1998)LeCun, Bottou, Bengio, and Haffner]{yann}
LeCun, Yann, Bottou, Leon, Bengio, Yoshua, and Haffner, Patrick.
\newblock Gradient-based learning applied to document recognition.
\newblock 1998.
\newblock URL \url{http://yann.lecun.com/exdb/publis/pdf/lecun-01a.pdf}.

\bibitem[medium(2017)]{relu}
medium.
\newblock relu operation.
\newblock 2017.
\newblock URL
  \url{https://medium.com/data-science-group-iitr/building-a-convolutional-neural-network\\-in-python-with-tensorflow-d251c3ca8117}.

\bibitem[Mobiny \& Nguyen(June, 2018)Mobiny and Nguyen]{lung_cancer}
Mobiny, Aryan and Nguyen, Hien~Van.
\newblock Fast capsnet for lung cancer screening.
\newblock June, 2018.
\newblock URL \url{https://arxiv.org/pdf/1806.07416.pdf}.

\bibitem[Mukhometzianov \& Carrillo(May, 2018)Mukhometzianov and
  Carrillo]{CapsNet_test_eval}
Mukhometzianov, Rinat and Carrillo, Juan.
\newblock Capsnet comparative performance evaluation for image classification.
\newblock May, 2018.
\newblock URL \url{https://arxiv.org/pdf/1805.11195.pdf}.

\bibitem[Pechyonkin(February, 2018)]{medium:capsnet}
Pechyonkin, Max.
\newblock Understanding hinton's capsule networks. part iv: Capsnet
  architecture.
\newblock February, 2018.
\newblock URL
  \url{https://medium.com/@pechyonkin/part-iv-capsnet-architecture-6a64422f7dce}.

\bibitem[Peng et~al.(2016)Peng, Wang, Chen, and Liu]{sequence}
Peng, Min, Wang, Chongyang, Chen, Tong, and Liu, Guangyuan.
\newblock Nirfacenet: A convolutional neural network for near-infrared face
  identification.
\newblock 2016.
\newblock URL \url{http://www.mdpi.com/2078-2489/7/4/61}.

\bibitem[Picasso(1923)]{picasso_face}
Picasso, Pablo.
\newblock Portrait de femme au col d`hermine.
\newblock 1923.
\newblock URL \url{https://www.wikiart.org/en/pablo-picasso/untitled-1937-8}.

\bibitem[quora(2016)]{conv}
quora.
\newblock convolutional operation.
\newblock 2016.
\newblock URL
  \url{https://www.quora.com/How-common-is-it-for-neural-networks\\-to-be-represented-by-3rd-order\\-tensors-or-greater}.

\bibitem[quora(2017)]{maxpool}
quora.
\newblock max pooling.
\newblock 2017.
\newblock URL
  \url{https://www.quora.com/What-is-max-pooling-in-\\convolutional-neural-networks}.

\bibitem[Ross(November, 2017)]{medium:visualpic}
Ross, Mike.
\newblock A visual representation of capsule network computations.
\newblock November, 2017.
\newblock URL
  \url{https://medium.com/@mike_ross/a-visual-representation-of-capsule\
  -network-computations-83767d79e737}.

\bibitem[Ruder(June, 2017)]{batch_gradient_descent}
Ruder, Sebastian.
\newblock An overview of gradient descent optimization algorithms.
\newblock June, 2017.
\newblock URL \url{https://arxiv.org/pdf/1609.04747.pdf}.

\bibitem[Shahroudnejad et~al.(February, 2018)Shahroudnejad, Mohammadi, and
  Plataniotis]{explainability}
Shahroudnejad, Atefeh, Mohammadi, Arash, and Plataniotis, Konstantinos~N.
\newblock Improved explainability of capsule networks: Relevance path by
  agreement.
\newblock February, 2018.
\newblock URL \url{https://arxiv.org/pdf/1802.10204.pdf}.

\bibitem[Timofte et~al.(2011)Timofte, Zimmermann, and van Gool]{belgium}
Timofte, Radu, Zimmermann, Karel, and van Gool, Luc.
\newblock \emph{Multi-view traffic sign detection, recognition, and 3D
  localisation}.
\newblock Journal of Machine Vision and Applications, Springer-Verlag, 2011.
\newblock \url{https://btsd.ethz.ch/shareddata/}.

\bibitem[University(2018)]{yale}
University, Yale.
\newblock Yale face database b.
\newblock 2018.
\newblock URL
  \url{http://cvc.cs.yale.edu/cvc/projects/yalefaces/yalefaces.html}.

\bibitem[Wikipedia(2018{\natexlab{a}})]{receptive_field}
Wikipedia.
\newblock Receptive field.
\newblock 2018{\natexlab{a}}.
\newblock URL
  \url{https://en.wikipedia.org/wiki/Receptive_field#Visual_system}.

\bibitem[Wikipedia(2018{\natexlab{b}})]{wiki:dotproduct}
Wikipedia.
\newblock Dot product.
\newblock 2018{\natexlab{b}}.
\newblock URL \url{https://en.wikipedia.org/wiki/Dot_product}.

\bibitem[Wikipedia(2018{\natexlab{c}})]{wiki_convnets}
Wikipedia.
\newblock Convolutional neural network.
\newblock 2018{\natexlab{c}}.
\newblock URL \url{https://en.wikipedia.org/wiki/Convolutional_neural_network}.

\bibitem[Wikipedia(2018{\natexlab{d}})]{wiki_mnist}
Wikipedia.
\newblock Mnist database.
\newblock 2018{\natexlab{d}}.
\newblock URL \url{https://en.wikipedia.org/wiki/MNIST_database#Performance}.

\bibitem[Wikipedia(2018{\natexlab{e}})]{wiki_softmax}
Wikipedia.
\newblock Softmax function.
\newblock 2018{\natexlab{e}}.
\newblock URL \url{https://en.wikipedia.org/wiki/Softmax_function}.

\bibitem[Zhang(2016)]{uniform_distribut}
Zhang, Zhifei.
\newblock Derivation of backpropagation in convolutional neural network (cnn).
\newblock 2016.
\newblock URL
  \url{https://pdfs.semanticscholar.org/5d79/11c93ddcb34cac088d99bd0cae9124e5dcd1.pdf}.

\end{thebibliography}
\bibliographystyle{icml2016}

\end{document}